\documentclass[lettersize,journal]{IEEEtran}
\usepackage{amsmath,amsfonts}
\usepackage{algorithm}
\usepackage{array}
\usepackage[caption=false,font=normalsize,labelfont=sf,textfont=sf]{subfig}
\usepackage{textcomp}
\usepackage{stfloats}
\usepackage{url}
\usepackage{verbatim}
\usepackage{graphicx}
\usepackage{cite}

\usepackage[T1]{fontenc}
\usepackage{amssymb}
\usepackage{algorithm}
\usepackage{algorithmicx}
\usepackage{algpseudocode}
\usepackage{graphicx}
\usepackage{booktabs}
\usepackage{color}
\usepackage{comment}
\usepackage{multicol}

\floatname{algorithm}{Algorithm}

\usepackage{amsthm}

\usepackage{multirow}

\begin{document}

\title{An Automated Reinforcement Learning Reward Design Framework with Large Language Model for Cooperative Platoon Coordination}

\author{Dixiao Wei, Peng Yi, Jinlong Lei, Yiguang Hong, Yuchuan Du
	\thanks{Dixiao Wei, Peng Yi, Jinlong Lei and Yiguang Hong are with  the Shanghai Research Institute for Intelligent Autonomous Systems, Tongji University, Shanghai, 201210, China, and Peng Yi, Jinlong Lei, Yiguang Hong are also with the Department of Control Science and Engineering, Tongji University,  Shanghai, 201804, China; 
    Yuchuan Du is with School of Transportation Engineering, and the
 Key Laboratory of Road and Traffic Engineering, Ministry of Education, Tongji University, Shanghai, 201804, China.
        E-mails: {\tt\small \{weidx, yipeng,leijinlong,yghong, ycdu\}@tongji.edu.cn}}
}


\maketitle

\begin{abstract}
Reinforcement Learning (RL) has demonstrated excellent decision-making potential in platoon coordination problems.
However, due to the variability of coordination goals, the complexity of the decision problem, and the time-consumption of trial-and-error in manual design, finding a well performance
reward function to guide RL training to solve complex platoon coordination problems remains challenging.
In this paper, we formally define the Platoon Coordination Reward Design Problem (PCRDP),  extending the RL-based cooperative platoon coordination problem to incorporate automated reward function generation. 
To address PCRDP, we propose a Large Language Model (LLM)-based Platoon coordination Reward Design (PCRD) framework, which systematically automates reward function discovery through LLM-driven initialization and iterative optimization.
In this method, LLM first initializes reward functions based on environment code and task requirements with an Analysis and Initial Reward (AIR) module, and then iteratively optimizes them based on training feedback with an evolutionary module.
The AIR module guides LLM to deepen their understanding of code and tasks through a chain of thought, effectively mitigating hallucination risks in code generation.
The evolutionary module fine-tunes and reconstructs the reward function, achieving a balance between exploration diversity and convergence stability for training.
To validate our approach, we establish six challenging coordination scenarios with varying complexity levels within the Yangtze River Delta transportation network simulation.
Comparative experimental results demonstrate that RL agents utilizing PCRD-generated reward functions consistently outperform human-engineered reward functions, achieving an average of 10\% higher performance metrics in all scenarios.
\end{abstract}


\section{Introduction}
With the development of connected and automated vehicle technologies, truck platooning technology has emerged in truck freight scenarios. 
Truck platooning refers to multiple trucks collaborating to travel in a line at short longitudinal distances. 
The truck platooning technology has been widely validated for its ability to reduce fuel consumption during long-distance travel \cite{jiang2024cooperative}, lower carbon dioxide emissions \cite{tsugawa2014results}, and enhance traffic efficiency. 
Increasingly, research is focusing on the platoon coordination problem, which involves coordinating the routes \cite{van2017fuel, xu2022optimizing,hu2024optimal}, speeds \cite{hoef2019predictive, liang2015heavy}, and departure times \cite{choi2024optimizing, bouchery2022coalition,bai2023large} of vehicles within a transportation network to form a platoon during travel, thereby achieving benefits. 
Larsson et al. \cite{larsson2015vehicle} demonstrated that the platoon coordination problem is an NP-hard combinatorial optimization problem. 
The diversity of freight tasks and coordination goals among vehicles in transportation scenarios further complicates the solution to this problem. 
Previous work has explored methods such as mixed integer programming \cite{van2017efficient}, dynamic programming\cite{bai2023large,xiong2024approximate}, model predictive control\cite{bai2021event}, and heuristics\cite{hoef2019predictive} for specific platoon coordination problems.
However, the performance and flexibility of manually designed algorithms are often limited by the problem size and varying traffic conditions.

Multi-Agent Deep Reinforcement Learning (MADRL) can make effective sequential decisions, thus addressing varying and large-scale traffic coordination problems \cite{pan2024towards,kumaravel2021optimal,pan2024heterogeneous}.
Wei et al. \cite{wei2023multi} first applied MADRL to the platoon coordination problem of planning departure times at hubs,  successfully training a strategy capable of handling multiple freight tasks. Currently, experts manually analyze the problem and design the corresponding reward function code for providing incremental learning signals in Reinforcement Learning (RL) training \cite{singh2009rewards,laud2004theory}. 
However, due to the  dynamics of traffic environment, as well as the various coordination goals and freight tasks, the objective functions and constraints of platoon coordination  can be highly complicated \cite{bouchery2022coalition, johansson2023hub,hu2024optimal}. 
This complexity poses significant challenges for traditional reward design approaches, as manually designing reward functions needs time-consuming trial-and-error.
In a recent experiment on expert-designed reward functions \cite{booth2023perils}, 83\% of experts selected valid reward functions in easy tasks, but only 47\% of them are able to designe effective reward functions under multiple task objectives. 
These results highlight the increased difficulty for human experts to manually design high-quality reward functions as the complexity of platoon coordination problems grows.

Recently, Large Language Models (LLMs) have become active in the field of reinforcement learning reward function code generation \cite{yu2023language,maeureka,li2024auto,xietext2reward,han2024autoreward}, due to their ability to understand scenarios, reason task objectives and analyze data. 
By inputting textual or code descriptions of scenarios and tasks into the LLM, it can generate a reasonable reward function code directly. 
However, the hallucination problem \cite{ji2023towards,liu2024exploring,yao2023llm} can lead to an inefficient or incorrect reward function code with a single query. 
By incorporating evolutionary ideas, Ma et al.\cite{maeureka} generated multiple candidate reward function codes to train repeatedly for mitigating the catastrophic effects of hallucinations, which increases training costs and token consumption. 
Sun et al. \cite{sun2024large} introduced trajectory preference evaluation to filter out low-quality reward function codes and reduce token consumption, but performance degradation occurred in the later stages of the evolutionary search for reward function codes.
Mitigating the impact of LLM illusion on reward function code generation and improving LLM's search ability for reward function code remains a huge challenge in LLM-based reward design.

\begin{figure*}[t]
    \centering
    \includegraphics[scale=0.10]{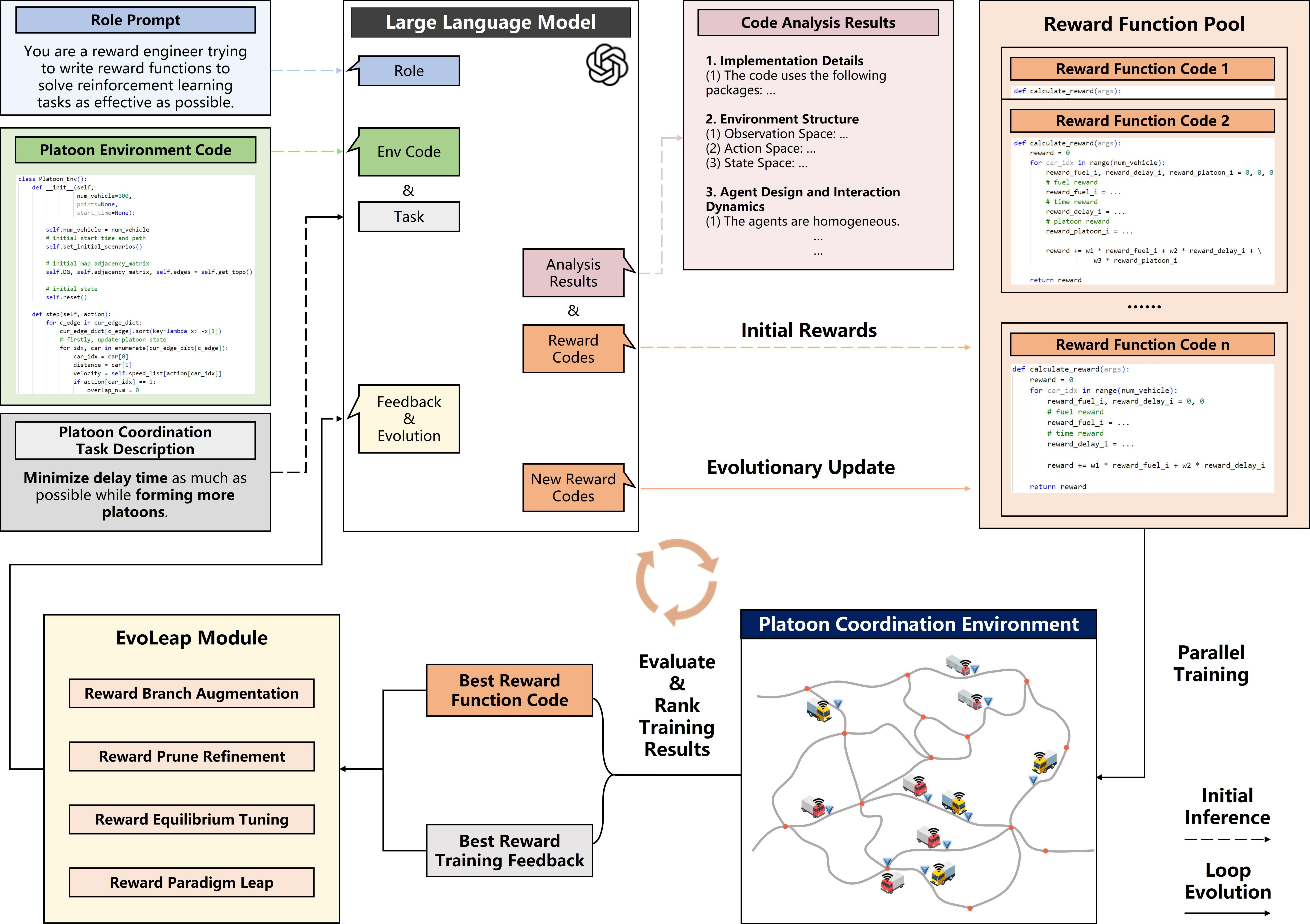}
    \caption{The working flow of PCRD. The steps in the dashed process demonstrate that LLM first analyzes the environment code and task requirements, and then initializes the reward function code. Then, the solid flow represents the evolution process of the reward function code, including reinforcement learning training and LLM improvement.}
    \label{fig1}
\end{figure*}

To address the aforementioned issues, we propose an LLM-based Platoon Coordination Reward Design (PCRD) framework to automate the reward designing of RL training to solve platoon coordination problems. 
The PCRD framework is composed of three key components: the Analysis and Initial Reward (AIR) module, MADRL parallel training, and the evolutionary and leap (EvoLeap) module.
The AIR module first analyzes the composition of the environment code across multiple dimensions, guided by the Chain of Thought (CoT). 
Then it generates several initial reward function codes based on the task requirements. 
This enables the LLM to better understand the relationship between code variables and the environment's Markov process, reducing low-quality or incorrect code generation due to hallucinations. 
These initial reward function codes are stored in the reward function code pool as the search starting points for reinforcement learning training.
Subsequently, the system automatically conducts parallel training using the reward function codes from the pool. 
After training is complete, the training data are filtered and sorted to select the optimal reward function as the evolutionary template.
The EvoLeap module then evolves the template for the next training cycle, based on the training feedback. 
This module involves fine-tuning and leaping the reward function to balance search breadth and stability.
Through multiple evolution cycles, PCRD continuously searches and refines reward functions.
The contributions of this paper can be summarized as follows:
\begin{itemize}
    \item We formalize the cooperative platoon coordination problem under various tasks and goals, and delve into the expanded reward design problem within this context.
    \item We present PCRD, an automated reward design framework that tackles platoon coordination challenges. It can analyze task requirements and environmental code for different platoon coordination problems, automatically searching and evolving reward functions to aid reinforcement learning training.
    \item We design the AIR and EvoLeap modules to boost the framework's search efficiency. 
     The AIR module uses the chain of thought to guide in-depth environmental code analysis by LLMs, reducing code error rates and enhancing the search starting point quality. 
     The EvoLeap module directs LLMs to fine-tune and innovate reward functions, stabilizing and improving the search direction and efficiency.
    \item We verify the framework's effectiveness by experimenting on  six scenarios of varying difficulty and coordination goals over the Yangtze River Delta transportation networks. 
    The results show that the reward functions designed by PCRD outperform those designed by human experts by 10\% in the six scenarios.
\end{itemize}

The remainder of the paper is organized as follows.
Section II summarizes the research status of platoon coordination and LLM-related issues. 
Section III introduces the problem formulation.
Section IV introduces the details of PCRD.
Section V shows experiments in a large-scale Chinese transport network. 
Section VI concludes this paper and outlines future research directions.

\section{Related Work}
\subsection{Platoon Coordination}
Platoon coordination focuses on optimizing vehicle driving strategies to form platoons that improve fuel efficiency. This has gained significant attention over the past decade due to its potential benefits \cite{lesch2021overview}.
For example, \cite{larson2013coordinated} investigated the planning of vehicle paths to enable platoon formation. 
Building on this foundation, \cite{liang2013fuel,liang2015heavy} explored the adjustment of vehicle speeds along their routes to facilitate gradual convergence into platoons. 
Subsequently, \cite{van2017fuel,van2015fuel} extended these efforts by simultaneously optimizing both routes and speeds, which expanded the opportunities for platoon formation.
More recently, studies such as \cite{zhang2017freight,johansson2018multi,boysen2018identical,johansson2021strategic} introduced an innovative strategy that schedules vehicle departure times at hubs, allowing vehicles with identical departure times and destinations to automatically form platoons.
Furthermore, the scope of platoon coordination has evolved to embrace multi-objective optimization, incorporating factors such as delay time \cite{johansson2022platoon}, traffic flow \cite{choi2024optimizing}, platoon size \cite{hu2024optimal}, and driver rest periods \cite{xu2022truck}. 
As research in this field progresses, the platoon coordination problems under investigation have become increasingly practical and aligned with real-world complexities, though this has also heightened their intricacy.

However, those traditional methods face significant computational complexity with large-scale scenarios \cite{larson2013coordinated, liang2013fuel, van2015fuel} and communication issues such as latency \cite{zhang2017platoon}. 
To address these limitations, the research community has been investigating AI-based approaches to enhance distributed and autonomous platoon coordination.
For example, Hoef et al. \cite{hoef2019predictive} proposed a distributed framework that employs heuristic algorithms to generate efficient formation plans for large-scale transportation networks.
Similarly, Bai et al. \cite{bai2021event} introduced a distributed model predictive control framework to optimize truck waiting times at hubs. 
In pursuit of rapid solutions, RL and approximate dynamic programming are recently explored   for platoon coordination. 
Bai et al. \cite{bai2023large} designed stage and terminal reward functions within an approximate dynamic programming framework to tackle the complexities of large-scale multi-fleet platoon coordination. 
Meanwhile, Xiong et al. \cite{xiong2024approximate} incorporated a reward function based on fuel efficiency and time considerations into their approximate dynamic programming method, aiming to improve coordination in mixed traffic flows comprising both autonomous and non-autonomous vehicles.
Wei et al. \cite{wei2023multi} utilized RL algorithms to solve large-scale platoon coordination problems under partial information. Their method balanced truck queues and waiting times at hubs to form effective platoons, thereby improving the efficiency of the transportation network. 
Nevertheless, they all developed a dense reward function for RL training by decomposing the objective function of platoon coordination. 

Despite these advancements, a critical drawback persists in approximation and learning-based methods: they rely on manually crafted reward functions tailored to specific problem scenarios. This manual design process is not only time-consuming but also constrains the adaptability of these approaches across diverse contexts. To overcome this limitation, our work proposes an automated framework for designing reward functions for platoon coordination problems. This framework is intended to efficiently address a wide range of coordination scenarios, enhancing both flexibility and scalability.

\subsection{Reward Design}
Reward design or credit assignment is a crucial aspect of reinforcement learning and decision-making systems, as it directly influences the behavior and performance of training autonomous agents. 
Over the years, various approaches and techniques have been explored to address the reward design problem.

Initially, reward functions were often designed manually based on domain knowledge and task requirements \cite{singh2009rewards,laud2004theory}. 
Experts would define specific reward values for different states and actions to guide the learning process of the agent \cite{chu2019multi,gupta2022unpacking}. 
For example, in robotics tasks, rewards might be assigned based on proximity to a target \cite{kober2013reinforcement}, energy consumption \cite{arulkumaran2017deep}, or the successful completion of a sub-task \cite{singh2022reinforcement}. 
This approach relies heavily on human experience and understanding of the problem domain. 
Although this method can be effective in well-defined scenarios, it becomes challenging when dealing with complex and dynamic environments where the optimal reward structure is not immediately apparent or sparse \cite{eschmann2021reward,ibrahim2024comprehensive}.

Inverse Reinforcement Learning (IRL) aims to learn the reward function by observing expert demonstrations rather than relying solely on manually designed rewards \cite{ng2000algorithms}. 
The underlying assumption is that the expert's behavior is optimal concerning an unknown reward function, which the algorithm seeks to recover \cite{abbeel2004apprenticeship}. 
Early IRL methods focused on maximum margin approaches \cite{ratliff2006maximum} and apprenticeship learning \cite{syed2007game}, where the learned policy should not be worse than the expert's policy. 
Subsequent advancements introduced maximum entropy principles \cite{ziebart2008maximum} to model the expert's behavior more accurately. 
These methods have been successfully applied in various domains \cite{levine2018learning,shah2022inverse,uchibe2018model}, including autonomous driving \cite{codevilla2018end}, where the agent learns to navigate by imitating human drivers' behavior.
However, IRL typically requires expensive expert data collection and outputs unexplainable black-box reward functions.

Recent work has considered using LLM to generate structured reward functions \cite{yu2023language,maeureka,li2024auto,xietext2reward,han2024autoreward}. 
Thanks to the vast amount of knowledge carried by LLM, designed rewards can become white boxes without the need for expensive manual trial-and-error or expert data.
Yu et al. \cite{yu2023language} connected low-level controllers to drive robots to complete simple tasks by designing vertical reward function interfaces.
Xie et al. \cite{xietext2reward} designed an iterative framework for reward functions based on human feedback. In this framework, LLM continuously modifies the reward function based on language feedback from human experts.
Furthermore, Ma et al. \cite{maeureka} proposed an automated reward function evolution framework that uses LLM instead of human experts to analyze training results as feedback and optimize the reward function.
Similar evolutionary methods have been applied to autonomous driving scenarios \cite{han2024autoreward}, and for complex driving domains, they propose environment code analysis modules and task structure modules to enhance LLM's understanding of scenarios and tasks.
Considering the improvement in LLM's ability to analyze training feedback, Li et al. \cite{li2024auto} utilized another LLM to summarize the reasons for failure based on policy trajectories to improve the reward function more effectively.
Sun et al. \cite{sun2024large} introduced trajectory preference evaluation to select high-quality reward functions, reducing token consumption.

However, these methods are primarily focused on common robot control tasks and analyze the success rate of a single task as feedback.
In specialized domains such as platoon coordination, these methods may suffer from hallucinations due to an insufficient understanding of the specific domain and complex objectives, leading to the generation of low-quality code and inefficient search.
In contrast, PCRD is designed to leverage LLM for the automatic design and evolution of reward functions for complex platoon coordination problems. 
The AIR module assists the LLM in understanding these intricate coordination issues, thereby enhancing the quality of the generated code. 
The EvoLeap module stabilizes the search process by constraining the direction of evolution.

\section{Problem Setting}
The goal of reward design is to design a state evaluation for optimizing a long-term objective function in a sequential decision-making problem. 
In this section, we first formulate the Cooperative Platoon Coordination Problem (CPCP) and reformulate it into a Decentralized-Partial Observable Markov Decision Process (Dec-POMDP). 
Then, we define the Platoon Coordination Reward Design Problem (PCRDP), motivated by the Reward Design Problem (RDP) in \cite{singh2009rewards}.

\subsection{Cooperative Platoon Coordination Problem}
We are considering a transportation network in which a set of vehicles is willing to cooperatively form a platoon. 
Mathematically, CPCP can be summarized as a tuple $(\mathcal{G},\mathcal{N},\mathcal{I},\mathcal{A},J,C)$.
\begin{itemize}
    \item [-] $\mathcal{G} = (\mathcal{V}, \mathcal{E})$ : A directed transportation graph consists of a hub set $\mathcal{V}$ and a edge set $\mathcal{E}$ that connect the hubs.
    \item [-] $\mathcal{N}=\{1,2,...,N\}$ : A set of trucks with freight tasks in the transportation network.
    \item [-] $\mathcal{I} = \{I_1,I_2,...,I_N\}$ : A set of trucks' freight information. For truck $i\in \mathcal{N}$, $I_i=(v_i^s,v_i^d,t_i^s,t_i^d)$ consists of a start hub $v_i^s \in \mathcal{V}$ at which truck $i$ will start its freight task at time $t_i^s$ and a destination $v_i^d \in \mathcal{V}$ at which truck $i$ has to finish its freight task before the deadline $t_i^d$.
    \item [-] $\mathcal{A}$ : Truck decision variable space. It may include traveling speed and hub departure time.
    \item [-] $J$ : Objective function. The goal of platoon coordination is mainly to maximize the fuel consumption saving after forming platoons \cite{davila2013environmental,bishop2017evaluation}. In addition, the objective function can also include penalties for waiting time \cite{johansson2023hub} and maintenance costs \cite{bouchery2022coalition}.
    \item [-] $C$ : Problem constraints. Constraints are often caused by practical factors, including arrival time constraints \cite{bai2022approximate}, communication constraints \cite{wei2023multi}, travel speed constraints \cite{hoef2019predictive}, platoon size constraints \cite{hu2024optimal}, rest rule constraints \cite{xu2022truck}, and traffic road constraints \cite{choi2024optimizing}.
\end{itemize}
CPCP aims at finding a set of decision sequence $X=\{x_1,x_2,...,x_N\}$ to maximize or minimize the objective function $J$:
\begin{equation}
    \begin{array}{ll}
    \max\limits_{x_i \in \mathcal{A},i \in \mathcal{N}}  J(X) \\
    \text { subject to } \mathcal{G},C,\mathcal{I}.
    \end{array}
\end{equation}
Here, $\forall i \in \mathcal{N}$, $x_i$ represents the decision sequence of the truck $i$ during the travel.
For a specific formulation, the reader can refer to \cite{wei2024multi}.

\subsection{Dec-POMDP Formulation for Cooperative Platoon Coordination}
Resorted to the MADRL method, we reformulated the CPCP as a Dec-POMDP \cite{oliehoek2016concise}. This Dec-POMDP is composed of a tuple $E_p = (\mathcal{N} , S, A, T, Z, O, R)$.
\begin{itemize}
    \item [-] $\mathcal{N}$ : A set of trucks with freight tasks in the transportation network.
    \item [-] $S$: The entire state of the transportation system. It may include the position, velocity, and delay time of each truck.
    \item [-] $A=\{a_1,a_2,...,a_N\}$: The joint action set of all trucks. Action $a_i$ may include the travel speed, the departure time of the hub, and the route of the freight, but may be restricted by the deadline, the driving rule, and the traffic situation.
    \item [-] $T:S \times A \times S \rightarrow [0,1]$: The state transition probability function. $T(s'|s,a)$ is the probability of transitioning from state $s$ to $s'$ with action $a$. This function describes the dynamic interaction process between trucks and the traffic environment.
    \item [-] $Z=\{z_1,z_2,...,z_N\}$: A set of trucks' observation. Observation $z_i$ represents the information that truck $i$ can collect while traveling in the transportation network. It may include the position, velocity, and delay time of the ego truck and nearby trucks, route information, and traffic situation.
    \item [-] $O$: Observation function. $z_i = O_i(S)$ represents truck $i$ can obtain partial information from the environment state $S$ due to possible communication limitations.
    \item [-] $R: S \times A \rightarrow \mathbb{R}$ : The reward function. $R(s,a)$ describes a scalar value when taking action $a$ in state $s$. In CPCP, the reward function $R$ should encourage the formation of platoons to optimize the objective function $J$.
\end{itemize}
The goal of Dec-POMDP is to train an optimal policy $\pi$ that maximizes the reward function $R$, and thus to optimize the objective function $J$ of CPCP:
\begin{equation}
    \mathbb{E}[\sum_{t=1}^{T_e}R(S^t,A^t)|\pi],
\end{equation}
where $T_e$ represents a maximal length allowing the last truck to reach its ending point, $S^t$ and $A^t$ are the state set and action set at time $t$.
Specifically, $\forall i \in \mathcal{N}$, $x_i=\{a_i^1,a_i^2,...,a_i^{T_e}\}$.
And $\forall t \in \{1,2,...,T_e\}$, $a_i^t \sim \pi(a_i|z_i^t)$.

\subsection{Reward Design Problem}
An RDP is composed of a tuple $(E,\mathcal{R},\Pi_E, F)$. 
\begin{itemize}
    \item [-] $E$: $E=(S,A,T)$ represents an actual environment model with state space $S$, action space $A$, and transition function $T$.
    \item [-] $\mathcal{R}$: The reward function space that maps an agent's state and action to a scalar primary reward $R \in \mathcal{R}$ that drives reinforcement learning. 
    \item [-] $\Pi_E$: The policy space. $\Omega_E(R):R\rightarrow \pi_E$ represents a reinforcement learning algorithm with a reward function $R$ that produces a policy $\pi_E \in \Pi_E$ in the Markov Decision Process (MDP), $(E,R)$. 
    \item [-] $F$: The fitness function. $F:\Pi_E\rightarrow \mathbb{R}$ is used to evaluate the performance of the policy in the environment. 
\end{itemize}
Thus, as shown in \eqref{eqrdp}, an RDP is to find a reward function $R$ such that a policy $\pi_E:= \Omega_E (R)$ is trained in the environment $E$ using a reinforcement learning algorithm to maximize the value of the fitness function $F(\pi_E)$.
\begin{equation}
    \label{eqrdp}
    \max \limits_{R \in \mathcal{R}} F(\Omega_E (R)).
\end{equation}

\subsection{Platoon Coordination Reward Generation Problem}
In PCRDP, we utilize LLM to search for an optimal reward function in the code space. 
We define the problem as a tuple $(E_p,\mathcal{R}_c,\Pi_{E_c}, J, J_{text}, \theta_L)$.
\begin{itemize}
    \item [-] $E_p$: The platoon coordination environment code. The detailed model is defined in Section 3.2.
    \item [-] $\mathcal{R}_c$: The reward function code space. The reward function $R_c \in \mathcal{R}_c$ is characterized in the form of code to guide reinforcement learning training.
    \item [-] $\Pi_{E_p}$: The policy space for CPCP. $\Omega_{E_p}(R_c):R_c\rightarrow \pi_{R_c}$ represents a reinforcement learning method that utilizes a reward function code $R_c$ to train a policy $\pi_{R_c} \in \Pi_{E_p}$ in the platoon coordination environment $E_p$. 
    \item [-] $J$: Fitness function $J:\Pi_{E_p}\rightarrow\mathbb{R}$ quantifying policy quality. It is equivalent to the objective function in CPCP.
    \item [-] $J_{text}$: Textual specification of $J$. 
    \item [-] $\theta_L$: The large language model. LLM can generate a reward code via the platoon coordination environment code and textual objective. $R_c \sim \theta_L(E_p, J_{text})$.

\end{itemize}
Thus, the platoon coordination reward generation problem aims to utilize LLM to search for a reward function code $R_c$ that maximizes the fitness function.
\begin{equation}
    \label{eqpcrdp}
    \max \limits_{R_c \in \mathcal{R}_c} J(\Omega_{E_p} (R_c)).
\end{equation}

\renewcommand{\thealgorithm}{1} 
    \begin{algorithm}
        \caption{Platoon Coordination Reward Design} 
        \begin{algorithmic}[1] 
            \Require Large language model $\theta_L$, platoon coordination environment code $E_p$, task description $J_{text}$, chat buffer $\mathcal{B}$, analysis prompt $P_A$, evolution prompts $P_{M}$, feedback prompt $P_{f}$, fitness function $J$, reinforcement learning method $\Omega_{E_p}$, iteration number $N_{iter}$, reward function code pool $\textbf{R}_c$.

            \State $\textbf{R}_c \leftarrow \emptyset$.
            \State $\mathcal{B}= \{E_p,J_{text},P_A\}$.
            \State // Generate $k$ basic reward function and store to $\textbf{R}_c$.
            \State $\{R_{c,1},R_{c,2},...,R_{c,k}\} \sim \theta_L(\mathcal{B})$.
            \State $\textbf{R}_c \leftarrow \{R_{c,1},R_{c,2},...,R_{c,k}\}$.
            \For {$N_{iter}$ iteration}
                \State // Training with each reward function in the pool.
                \State $\{\pi_{R_{c,1}},\pi_{R_{c,2}},...,\pi_{R_{c,|\textbf{R}_c|}}\} = \Omega_{E_p}(\textbf{R}_c)$.
                \State // Evaluate for each policy.
                \State $\{J_1,J_2,...,J_{|\textbf{R}_c|}\} = J(\{\pi_{R_{c,1}},\pi_{R_{c,2}},...,\pi_{R_{c,|\textbf{R}_c|}}\})$.
                \State Find the index of current best reward $i^*$ according (8) and (9).
                \State $R_b = R_{c,i^*}$.
                \State // Update reward function code pool.
                \State $\textbf{R}_c \leftarrow R_{b}$
                \For {evolution prompt $P_m$ in $P_M$}
                    \State $\mathcal{B}_m \leftarrow \mathcal{B} \cup \{R_{b}, P_{f}, J_{i^*}, P_m\}$.
                    \State // Reward evolution.
                    \State $\{R_{c,1},R_{c,2},...,R_{c,m}\} \sim \theta_L(\mathcal{B}_m)$
                    \State $\textbf{R}_c \leftarrow \textbf{R}_c \cup \{R_{c,1},R_{c,2},...,R_{c,m}\}$.
                \EndFor
            \EndFor
            \Ensure Best reward function $R_{b}$.
        \end{algorithmic}
    \end{algorithm}

\section{Method}
In this section, we provide a detailed introduction to the structure of PCRD.
As shown in Fig. \ref{fig1}, PCRD consists of three components: 
1) First, LLM analyzes the environment code from several aspects to provide richer information for reward code generation. Then, based on the analysis of the code content and the description of the truck formation task input, generate multiple reward function codes for the current task and store them in the reward function pool.
2) Automatic parallel training reward function pool through multi-agent deep reinforcement learning method.
3) Four concise prompts for improving the current most effective reward function code.
The proposed framework applies to multi-agent reinforcement learning training driven by any simulation environment. 
We will introduce each part of the framework in the following sections.

\begin{figure*}[t]
    \centering
    \includegraphics[scale=0.10]{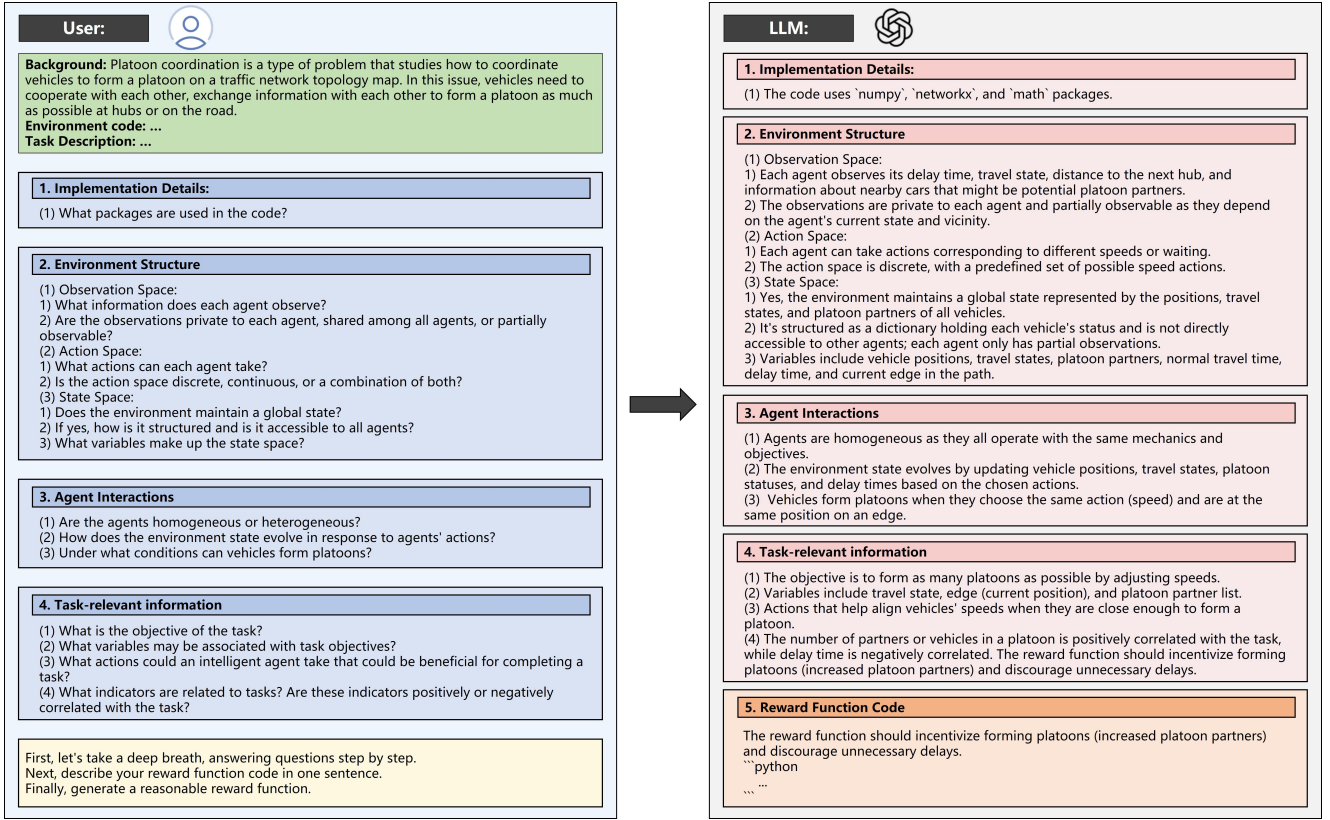}
    \caption{Implementation details of an AIR module example.}
    \label{figtalk}
\end{figure*}

\subsection{ Task Analysis and Initialize Reward}
The initial phase requires ensuring that large language models can comprehend complex environmental task code and generate accurate reward functions, which establishes a favorable foundation for subsequent search processes. 
Leveraging LLMs' exceptional contextual understanding capabilities, we jointly input the platoon coordination simulation environment code with textual task descriptions to facilitate reward function generation.
The simulation code encapsulates complete Dec-POMDP information, including environmental states, agent actions, observation functions, and state transition functions. 
This comprehensive information enables LLMs to accurately understand the platoon coordination problem and inherent relationships among variables. 
To ensure a fair comparison with human-designed reward functions, all variables and functions associated with expert implementations were deliberately obscured.

However, the highly specialized nature of CPCPs and the inherent complexity of environmental codes present significant challenges. 
While the environment code provides technical specifications, it fails to explicitly convey crucial background knowledge - such as platoon formation conditions - essential for proper understanding. 
Preliminary experiments revealed that LLMs tend to generate hallucinations \cite{chang2024survey,han2024autoreward} when interpreting the environment code, resulting in suboptimal reward functions characterized by logical inconsistencies or syntactical errors. 
Typical issues include referencing nonexistent environmental variables or implementing calculation logic incompatible with actual task requirements.

To address these limitations, we developed a chain-of-thought-guided framework employing an analyze-before-generation methodology. The framework implementation follows three systematic steps: First, providing LLMs with essential background knowledge about platoon coordination. Subsequently, LLMs conduct structured code analysis through four critical dimensions: implementation details, environmental architecture, agent interactions, and task-relevant information. Finally, explicitly specifying the fundamental structure of reward functions in prompts to enable LLMs to generate syntactically correct and logically coherent reward functions based on code analysis outcomes.

CoT is an effective technique that boosts the model's understanding and reasoning in a specific problem-solving direction, which in turn allows it to produce higher-quality outputs.
By guiding the LLM to answer the questions we have designed, we can enhance the LLM's ability to understand a specific problem direction, thereby generating better reward functions. 
Four-dimensional guidance details are shown as follows.
\begin{itemize}
    \item [-] \textbf{Implementation details:} The Q\&A on implementation details enables the LLM to clarify the packages used within the environment, reducing the errors of using other unintroduced packages. 
    \item [-] \textbf{Environmental architecture:} Answering questions about environmental architecture strengthens the LLM's understanding of the entire Dec-POMDP modeling information, clarifying information such as environmental states, observation functions, and agent actions. 
    \item [-] \textbf{Agent interactions:} Questions about agent interaction benefits the LLM in understanding interaction methods between multiple agents, such as understanding under what conditions vehicles can form a platoon in this environmental code.  
    \item [-] \textbf{Task-relevant information:} The Q\&A on task-related information allows the LLM to explicitly associate task information with variables in the environmental code, actively prompting the LLM to consider which variables and functions should be used to generate the reward function, reducing logical and syntactic errors in the code.  
\end{itemize}
Fig. \ref{figtalk} shows the guidance details of the AIR module.
In addition, we explicitly specify the basic structure of the reward function in the prompt so that the reward function can directly replace the original one. 
This guiding paradigm significantly reduces hallucinations of LLMs and can be applied to other similar environmental codes.

\subsection{Multi-Agent Reinforcement Learning Training and Evaluation}
The newly generated reward function code will be stored in the reward function pool for storage and used for training via multi-agent reinforcement learning methods. 
Since we have specified the basic format of the reward function, the new reward function code can easily overwrite the original reward function code. 
Subsequently, we train in parallel using these new reward functions. All the training processes are trained using the same multi-agent reinforcement learning algorithm and hyperparameters, with the only difference being the reward functions.
After each iteration of the training process, the proposed framework will automatically evaluate and filter these training results. 
We use $J$ as the evaluation metric in the training process to test the performance of these reward functions and select the current optimal reward function according to the filter rules.

Some works select the reward function with the maximum value of the evaluation metric among all training results as the filter rules. 
However, due to the non-stationarity of the platoon coordination multi-agent scenario and the composite indicators of CPCP, the training process is more prone to oscillations, falling into an inefficient local optimum, or premature degradation. 
Simply selecting the reward function with the maximum value as the filter rules can leave reward functions that do not converge in training as templates for subsequent improvement, leading to incorrect search directions. 
Therefore, we first filter out the training curves with better convergence effects and then select the reward function with the highest evaluation metric as the current optimal reward function for the following evolution.
Mathematically, for $n$ reward function codes $\{R_{c,1},R_{c,2},...,R_{c,n}\}$ and the training results $\{\mathcal{J}_1,\mathcal{J}_2,...,\mathcal{J}_n\}$, we determine their convergence and performance to find current best reward function code $R_{b}$ from the mean difference, standard deviation, and the overall slope trend.
Specifically, $\forall i \in \{1,2...,n\}$, $\mathcal{J}_i = \{J_i^1, J_i^2, ..., J_i^L\}$ is a fixed-length sequence $L$ from training. 

$\mathcal{F}_{mean}$ determines whether the evaluation metric has increased by calculating the comparison of the training results in the early and late stages.
\begin{equation}
\label{eqmean}
    \mathcal{F}_{mean} \triangleq \{i \mid J_i^{em} < J_i^{lm}, i\in \{1,...,n\}\}.
\end{equation}

Here, $J_i^{em}= \frac{1}{\alpha}\sum_{t=1}^\alpha J_i^t$ represents the mean in the early stage, $ J_i^{lm} = \frac{1}{\beta}\sum_{t=L-\beta+1}^L J_i^{t}$ is the mean in the late stage. $\alpha$ and $\beta$ represent the data length selected in the early and late stages.

$\mathcal{F}_{std}$ determines whether the training result has converged by calculating the variance ratio of the training results in the early and late stages.
\begin{equation}
\label{eqstd}
    \mathcal{F}_{std} \triangleq \{i \mid \frac{J_i^{lv}}{J_i^{ev}} < v_{th}, i\in \{1,...,n\}\}.
\end{equation}

Here, $J_i^{ev}= \sqrt{\frac{1}{\alpha}\sum_{t=1}^\alpha (J_i^t - J_i^{em})^2}$ represents the variance in the early stage, $ J_i^{lv} = \sqrt{\frac{1}{\beta}\sum_{t=L-\beta+1}^L (J_i^{t}-J_i^{lm})^2}$ is the variance in the late stage. $v_{th}$ is a volatility threshold.

$\mathcal{F}_{slope}$ uses linear regression to judge the overall trend of the training results.
\begin{equation}
\label{eqslope}
    \mathcal{F}_{slope} \triangleq \{i \mid \frac{J_i^{cov}}{J_i^{v}} > 0, i\in \{1,...,n\}\}.
\end{equation}

Here, $J_i^{cov} = \frac{1}{L}\sum_{t=1}^L (J_i^t - J_i^{m})(t-L^m)$ represents the covariance. 
$J_i^{m} = \frac{1}{L}\sum_{t=1}^L J_i^t$ and $L^m = \frac{1}{L}\sum_{t=1}^L t$ are the mean values of the abscissa and ordinate.
$J_i^{v} = \frac{1}{L}\sum_{t=1}^L (t-L^{m})^2$ is the variance.

Thus, we follow the above three functions to filter the training results:
\begin{equation}
    \mathcal{J}_{index} = \{i \mid \mathcal{F}_{mean} \cap \mathcal{F}_{std} \cap \mathcal{F}_{slope} \}.
\end{equation}
Then, the maximum metric of the training result will be chosen as the best current reward function code:
\begin{equation}
    i^*=\underset{i \in \mathcal{J}_{index}}{\arg\max}  \ (max(\mathcal{J}_i)),
\end{equation}
\begin{equation}
    R_{b} = R_{c,i^*}.
\end{equation}

\begin{figure*}[t]
    \centering
    \includegraphics[scale=0.11]{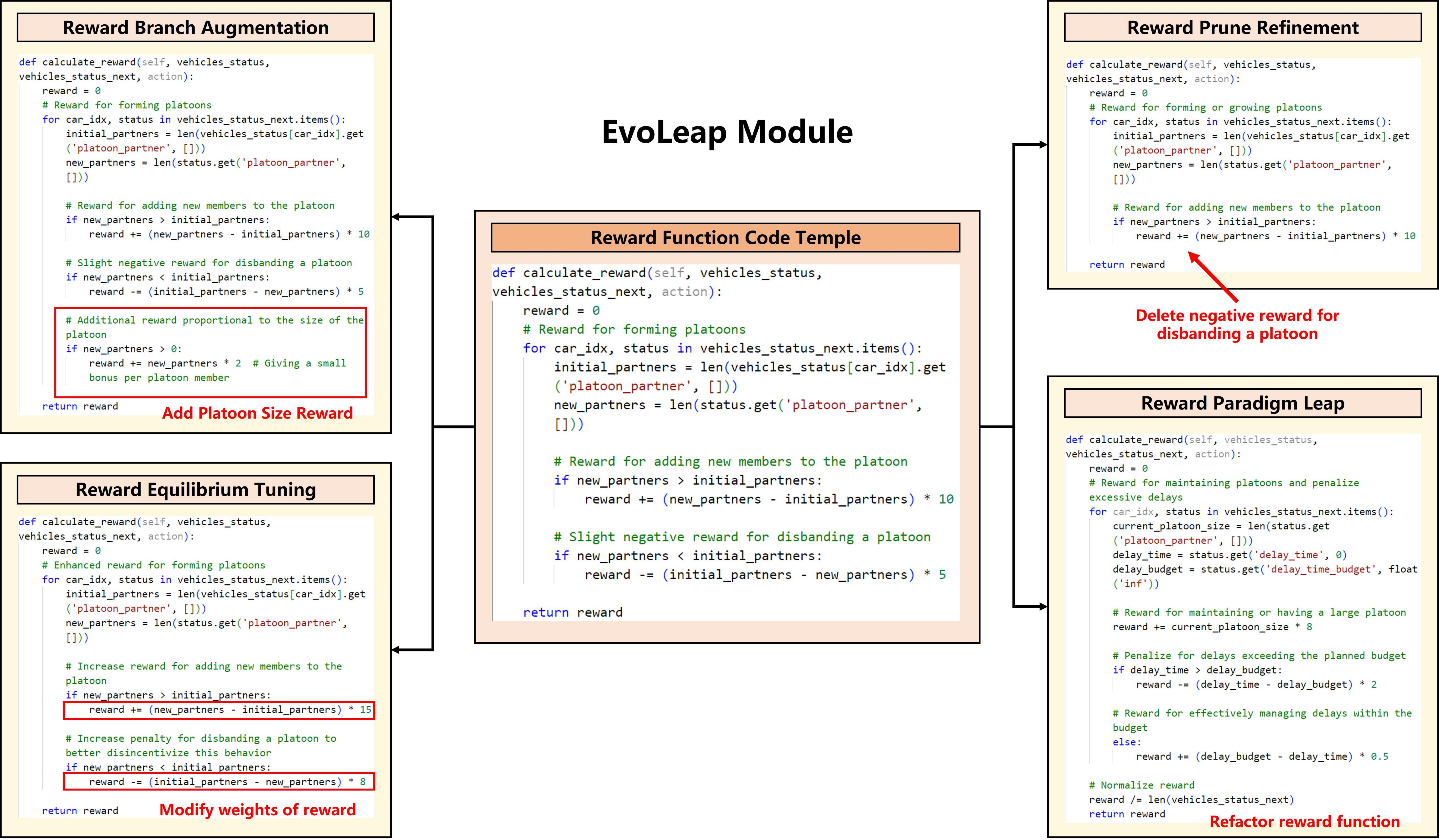}
    \caption{A case study of the EvoLeap module.}
    \label{figevolution}
\end{figure*}

\begin{table*}[h]
\centering
\caption{Evolutionary details of the EvoLeap module}
\begin{tabular}{cc}
\toprule
Evolutionary Strategy & Detail description                                                                                              \\
\cmidrule(r){1-2}
Fine-tuning 1 (F1)    & Reward Branch Augmentation. Add a new reward component while keeping the other parts of the function unchanged. \\
Fine-tuning 2 (F2)    & Reward Prune Refinement. Simplify the component of the original reward function.                                \\
Fine-tuning 3 (F3)    & Reward Equilibrium Tuning. Modify the weights of the reward components.                                         \\
Leap 1 (L1)           & Reward Paradigm Leap. Create a new reward function that has a different idea from the original reward function. \\ \bottomrule
\end{tabular}
\end{table*}

\subsection{Evolution Leap of Reward Function}
In practical engineering, human experts often use trial-and-error based on training feedback to refine the reward function \cite{booth2023perils}. 
This requires significant domain-specific expertise and manpower.
Hence, we propose an automated reward function evolution framework. 
It leverages LLMs to iteratively and intelligently search the reward function code space as per training feedback. 
The crux lies in devising an effective evolution approach to enhance the LLM's search efficiency and stability in this code space. 
Yet, given the vastness of the reward function code space and the LLM's context-based, probabilistic character-outputting mode, the generated reward functions may diverge much between iterations. 
Such an unstable search is likely to lead to degradation or an invalid reward function, thus reducing the search efficiency.
To address this, we design four concise improved prompts to steer the LLM towards refining the current best reward function.
They are divided into two categories: fine-tuning and leap. The fine-tuning strategy focuses on fine-tuning the original reward function to find a better reward function. The leap strategy tries to find a better reward function from other ideas.
Based on the current best reward function template and training feedback, LLM improves the reward function according to evolution prompts.
The specific form of training feedback is the curve of rewards and objective function values.
Fig. \ref{figevolution} shows a detail of using the EvoLeap module to evolve the reward function.
Details of these evolutions are shown in Tab. I.





As shown in Algorithm 1, PCRD first generates $k$ basic reward functions for exploration at the beginning. 
During evolution, each evolution strategy generates $m$ evolved reward functions. 
The maximum number of iteration rounds is set to $N_{iter}$. 
Thus, our algorithm needs to interact with the LLM $k + m(N_{iter}-1)$ times in total, with linear time complexity related only to $k$, $m$ and $N_{iter}$.
The number of iterations required to obtain a suitable reward function depends on the complexity of the platoon coordination task.
Furthermore, our method only needs to interact with the LLM before training. 
By optimizing the reward function through multiple brief interactions, the training strategy can perform better in complex platoon traffic environments.

\begin{figure}[h]
	\centering
	\includegraphics[scale=0.37]{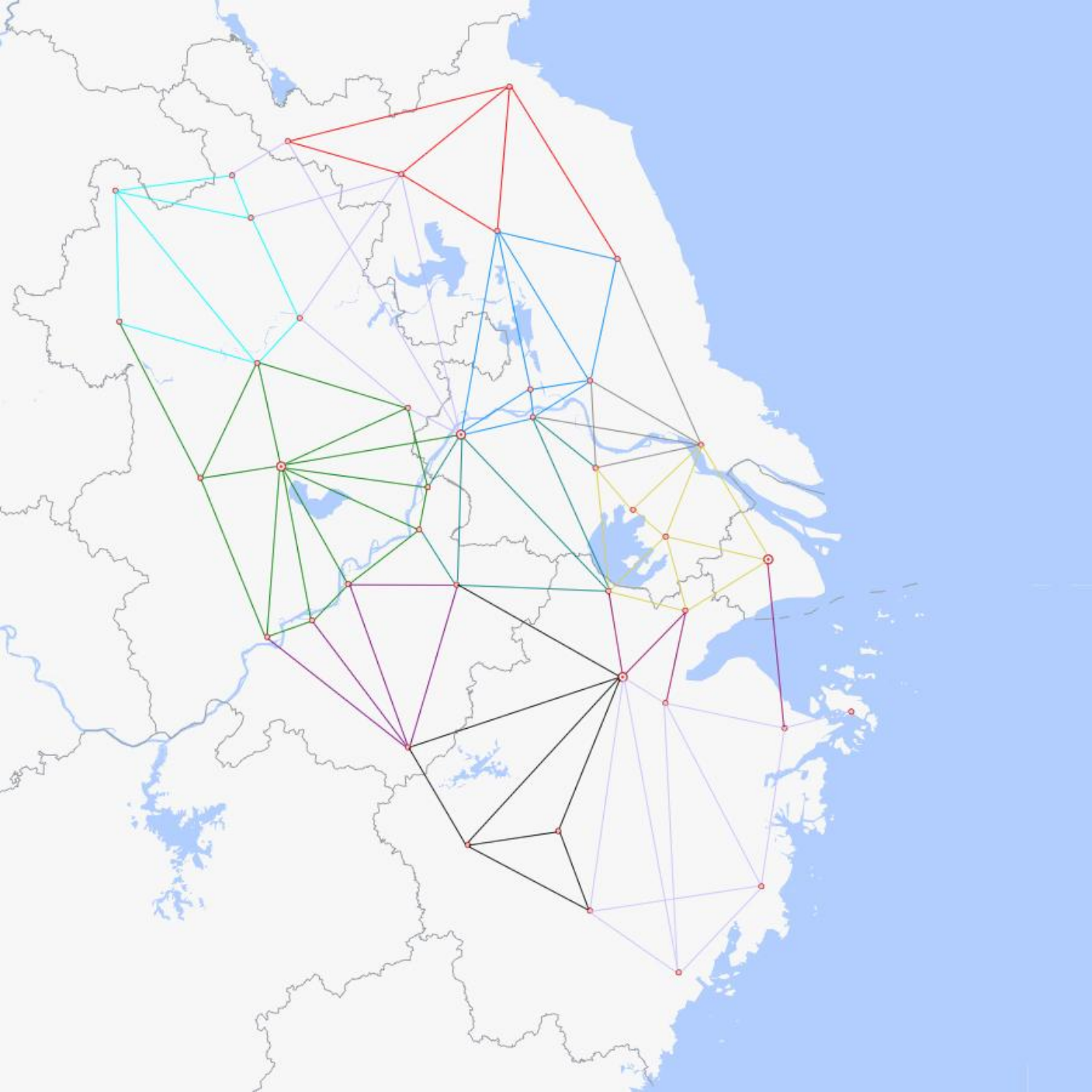}
	\caption{Transportation Network Map of Yangtze River Delta Region.}
	\label{figmap}
\end{figure}

\section{Experiments}
In this section, we conduct experiments on the transport network in the Yangtze River Delta area of China to demonstrate the effectiveness of our method.

\subsection{Experiment Setting}
Our experiments are implemented with an Intel(R) Xeon(R) Silver 4210R CPU @ 2.40GHz, 4 NVIDIA 4090 GPUs, and Pytorch.
As depicted in Fig. \ref{figmap}, we developed an agent-based simulation platform to model large-scale freight transportation systems within the Yangtze River Delta region. This simulation platform enables comprehensive scenario configuration through flexible control parameters including fleet size, departure/arrival schedules, route selection mechanisms, and optimization objectives.
The constructed network model comprises 41 interconnected nodes representing urban centers and 202 directional links. Geographically covering four provincial-level administrative divisions (Shanghai, Zhejiang, Jiangsu, and Anhui), the simulation infrastructure incorporates real-world spatial relationships derived from authoritative GIS data\footnote{Urban coordinates and road network parameters were acquired from www.amap.com/}. Each node corresponds to a metropolitan hub, while directed edges simulate intercity transportation corridors with bidirectional traffic capacity.

\begin{table*}[t]
\centering
\caption{Textual Objective function in six scenarios.}
\begin{tabular}{ccc}
\toprule
                         & Scenarios      & Textual specification $J_{text}$                                                                                                    \\
                         \cmidrule(r){1-3}
\multirow{3}{*}{Single-Object} & Wait  & Without considering time, planning the waiting time at hubs to form platoons as many as possible to obtain profits.    \\
                         & Speed & Without considering time, adjusting the speed to form platoons as many as possible to obtain profits.                  \\
                         & Mix   & Without considering time, adjusting the speed and waiting time to form platoons as many as possible to obtain profits. \\
                         \cmidrule(r){1-3}
\multirow{3}{*}{Multi-Object} & Wait  & Planning the waiting time at hubs to form as many platoons as possible while reducing delay time.                      \\
                         & Speed & Adjusting the speed to form as many platoons as possible while reducing delay time.                                    \\
                         & Mix   & Adjusting the speed and waiting time to form as many platoons as possible while reducing delay time.                  \\ \bottomrule
\end{tabular}
\end{table*}

\subsubsection{Simulation Setting}
Our simulation platform establishes a fleet size parameter \(|\mathcal{N}|=100\), where each truck agent follows a predetermined shortest route between origin-destination pairs.
The temporal parameters were randomized within a specified morning window (5:00 a.m.-11:00 a.m.), with discrete temporal resolution \(\Delta t = 5\) minutes and finite time horizon \(T_e=1000\) minutes. 
The simulation clock initialization aligns \(t=1\) with the baseline temporal coordinates at 5:00.
We implemented a hierarchical spatial decomposition of the network topology into 12 chromatic zones (Fig. \ref{figmap}), each containing 5-9 interconnected hubs. 
This zoned architecture constrains vehicle routing within designated sub-networks during training phases, deliberately increasing vehicular density to enhance platoon formation potential for improving the efficiency of training samples. 
We generated five distinct freight missions per zone (60 total freight missions), each specifying: geospatial constraints, including origin-destination nodal pairs; temporal distribution parameters, departure time sampling; path planning constraints, predefined optimal routes; and demographic-weighted routing probabilities, which are proportional to municipal population metrics \cite{johansson2021strategic}.

The simulation platform allows trucks to perform low-speed $v_l=60 km/h$, medium-speed $v_m=75 km/h$, and high-speed $v_h=90 km/h$ actions, and wait at hubs during their journeys. Given the sparse road network on the map, the fuel savings from platooning by changing routes are likely outweighed by the fuel consumption from detours. So, we assigned fixed shortest paths to each truck in the experiment. To avoid excessively long simulation times, we also set the deadline at the destination for each truck. Once a truck's delay caused by waiting or low speed exceeds the allowed time, it can no longer select low-speed or stationary actions until it accelerates to make up for the delay.
The maximum allowable delay time is set to 10\% of the time required for a truck to reach the destination at a constant speed of $v_m$.

\subsubsection{Scenarios Setting}
We designed six scenarios on the Yangtze River Delta transportation platform to verify the effectiveness and adaptability of our method. The scenarios included two different objective functions and three coordination methods.

\textbf{Single-Object:} The objective is to maximize the fuel profit of the platoon composed of truck drivers on the road. 
Within the allowable delay time, trucks can form as many platoons as possible by planning the speed or waiting time.
In this situation, the objective function does not consider the cost of time delay but only aims to form as many platoons as possible during the journey.
Like in \cite{wei2023multi}, the objective function can be defined as $J= c_p \times D_{p}$. 
$c_p=1.7$ represents platoon fuel profit per kilometer and $D_{p}$ is the distance by which the trucks travel in the platoon.

\textbf{Multi-Object:} The objective is to maximize the distance the trucks travel in the platoon while minimizing the delay time.
In this situation, the objective function is a multi-objective optimization problem considering both the cost of time delay and the benefits of the platoon.
According to \cite{wei2024multi}, the objective function is the difference between the platoon journey rate $P_j$ and the average delay time rate $T_r$: $J=P_j - T_r$.

$P_j$ represents the proportion of total travel distance completed in platoon formation, calculated as:
\begin{equation}
    P_j = \frac{D_p}{D_{total}}
\end{equation}
$D_{total}$ is the sum of distances traveled by all trucks.

$T_r$ quantifies the relative delay impact, expressed as:
\begin{equation}
    T_r = \frac{T_d}{T_{total}}
\end{equation}
$T_d$ represents the total delay time of all trucks. $T_{total}$ is the sum of the time required for the truck to reach the terminal at medium speed $v_m$.

These dimensionless parameters ($P_j$ and $T_r$) serve as normalized indicators for platooning efficiency and temporal cost respectively, enabling effective trade-off analysis between platoon benefits and scheduling constraints through their balanced combination in the objective function.

\begin{figure*}[t]

    \centering
    \subfloat[Single-Object Wait Scenario]{
    \includegraphics[scale=0.28]{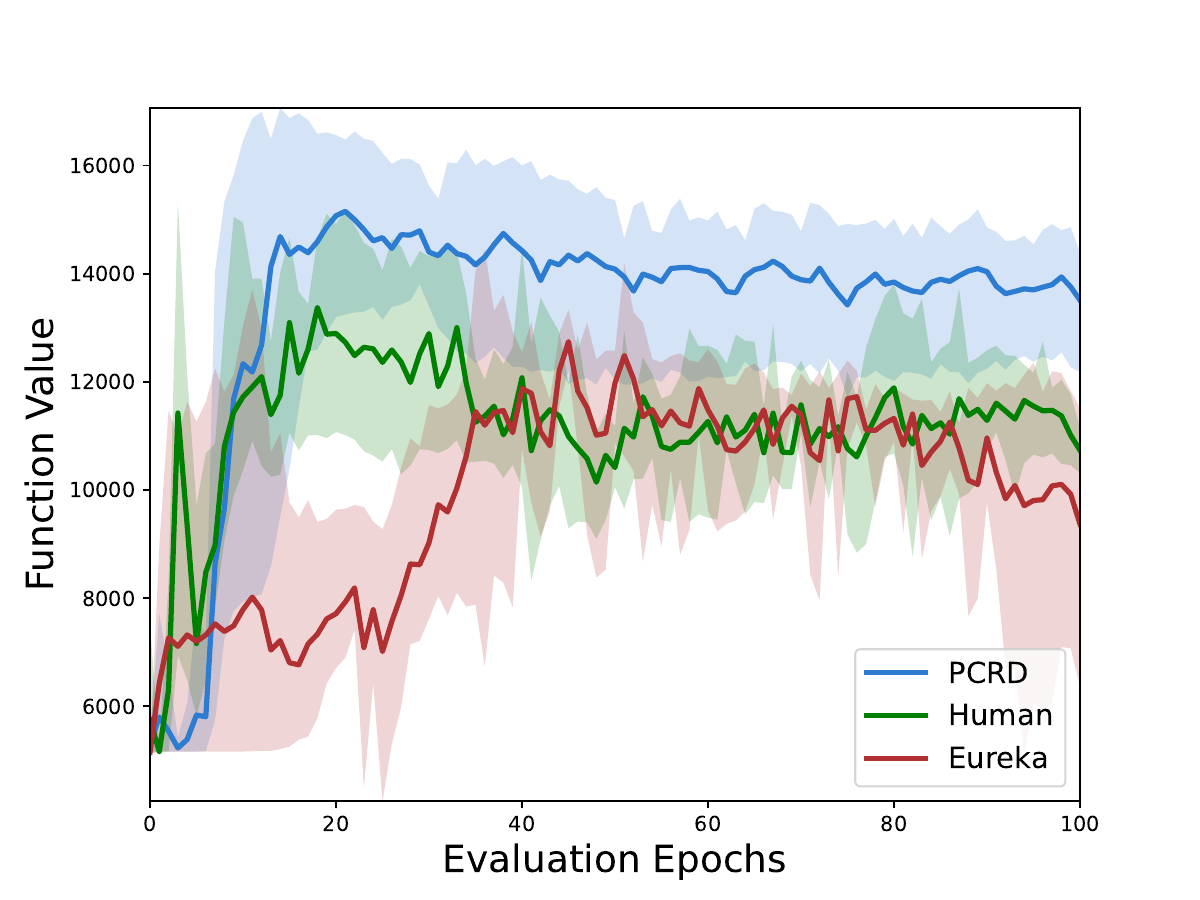}
    \label{figcomparisona}}
    \subfloat[Single-Object Speed Scenario]{
    \includegraphics[scale=0.28]{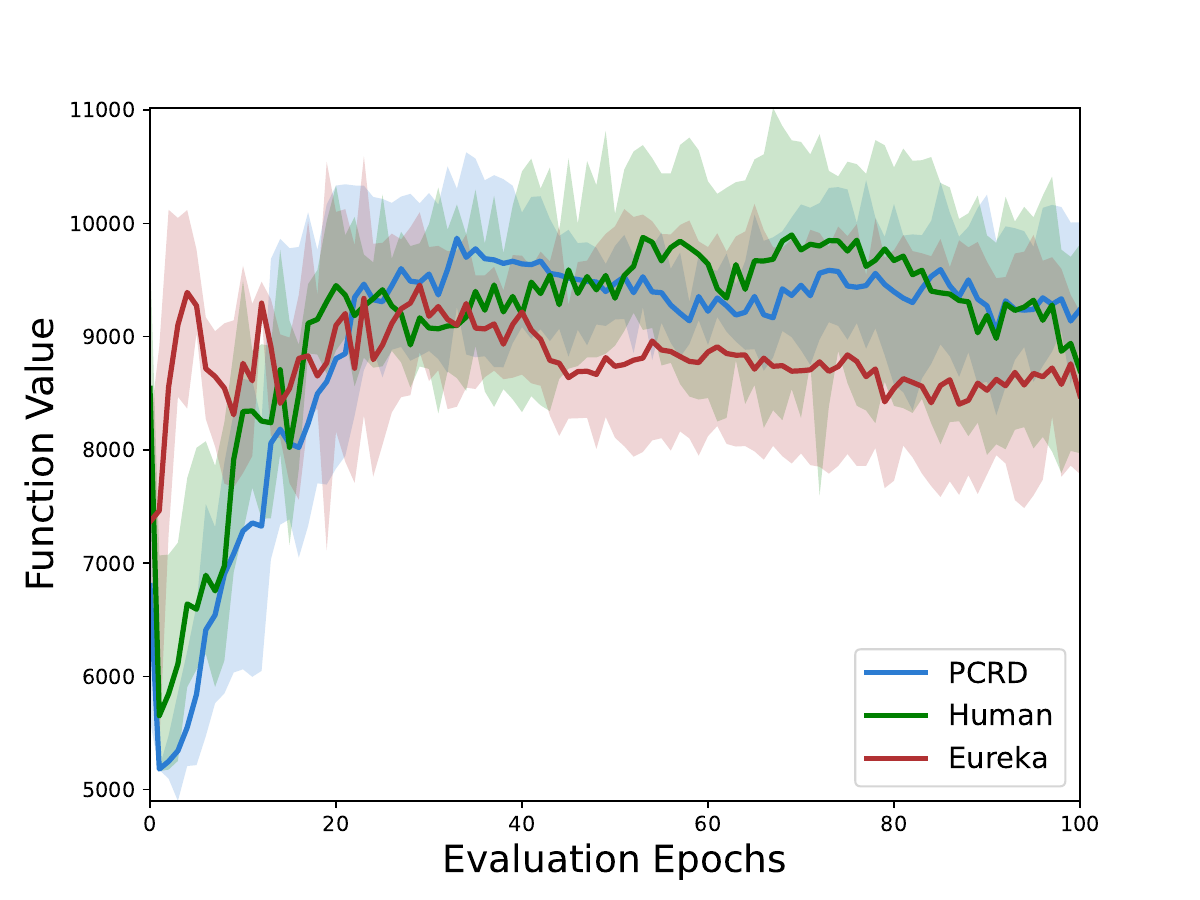}
    \label{figcomparisonb}}
    \subfloat[Single-Object Mix Scenario]{
    \includegraphics[scale=0.28]{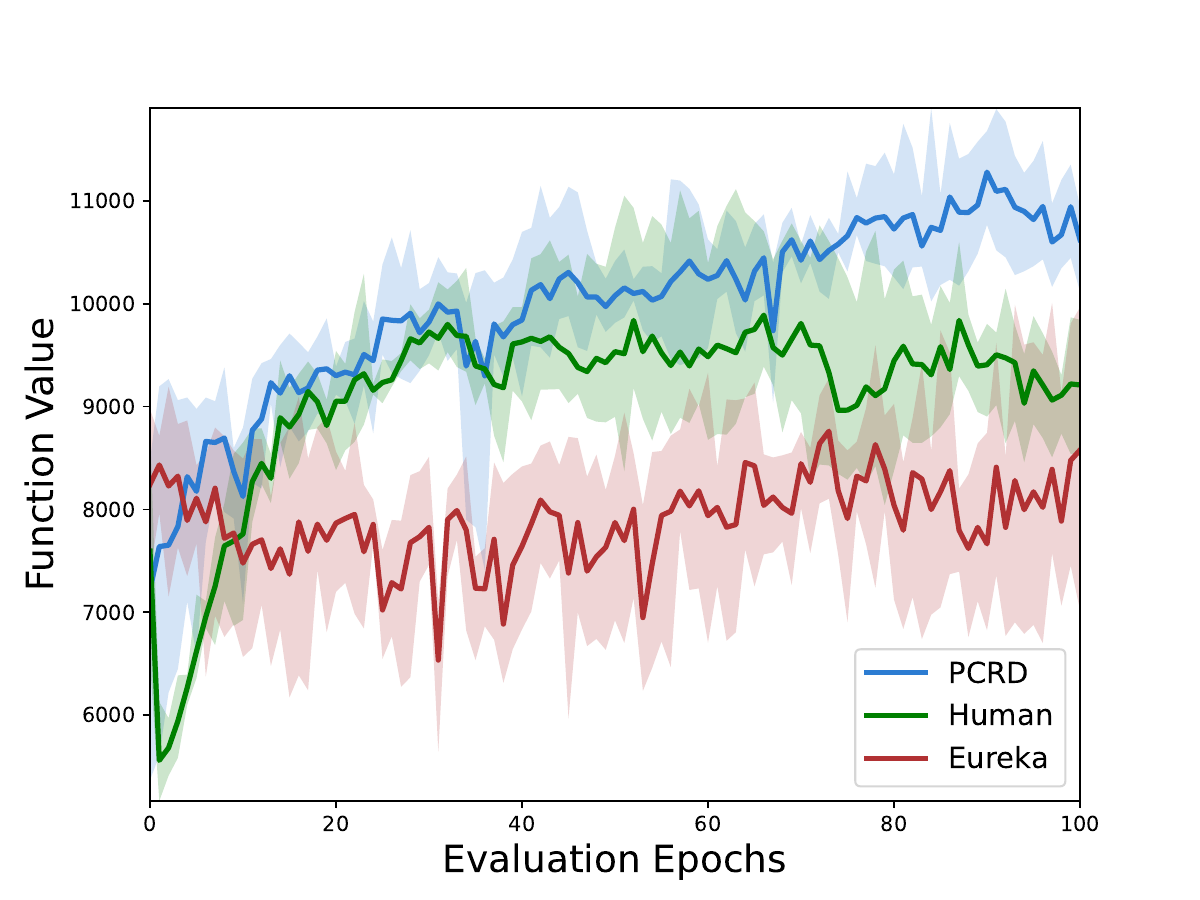}
    \label{figcomparisonc}} \\
    \subfloat[Multi-Object Wait Scenario]{
    \includegraphics[scale=0.28]{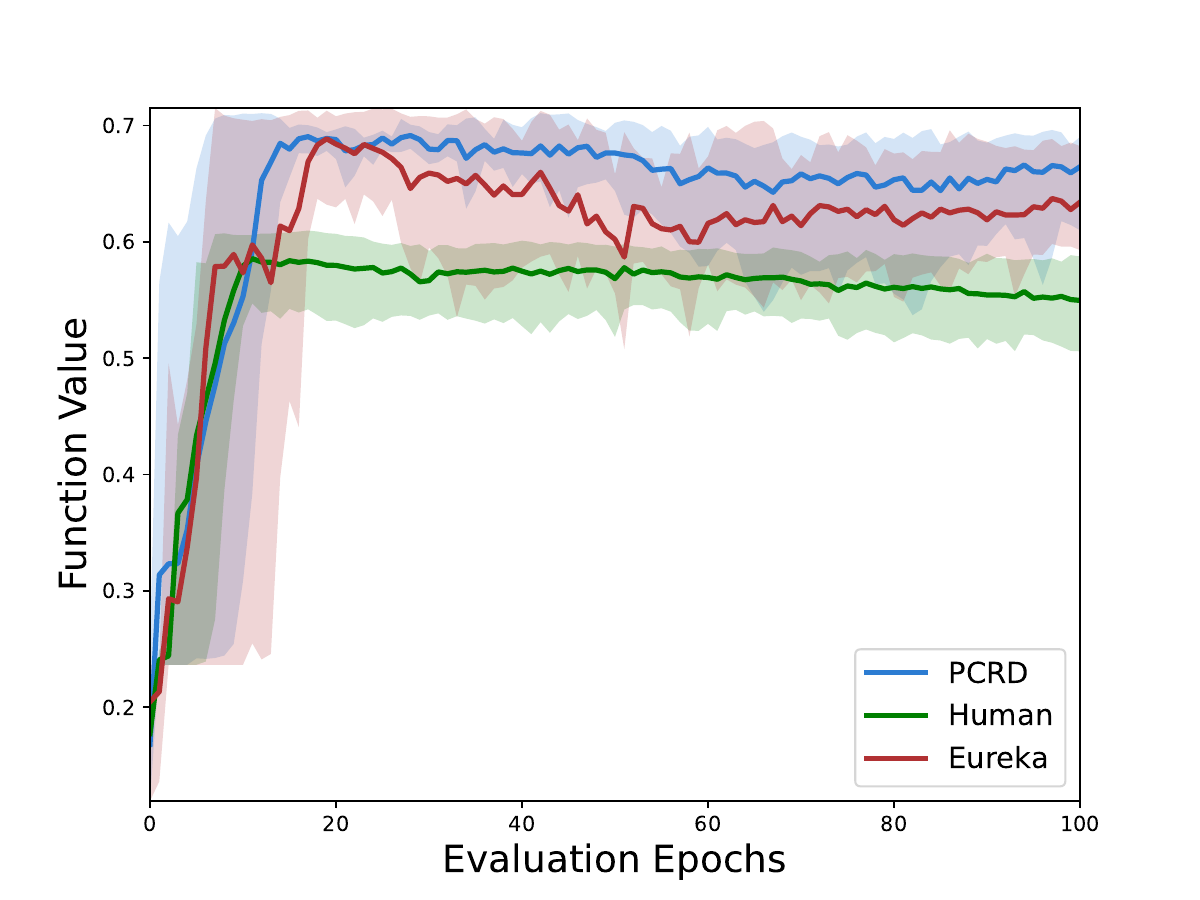}
    \label{figcomparisond}}
    \subfloat[Multi-Object Speed Scenario]{
    \includegraphics[scale=0.28]{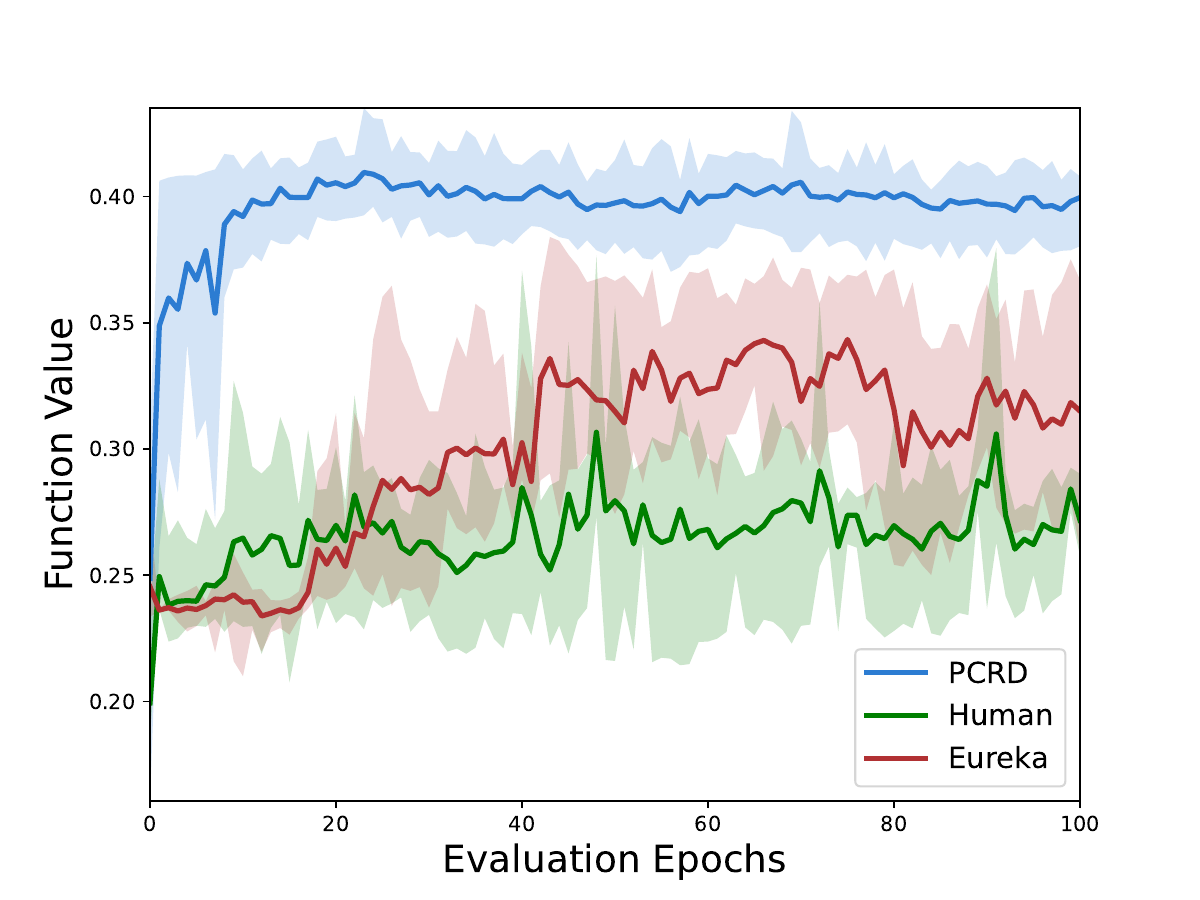}
    \label{figcomparisone}}
    \subfloat[Multi-Object Mix Scenario]{
    \includegraphics[scale=0.28]{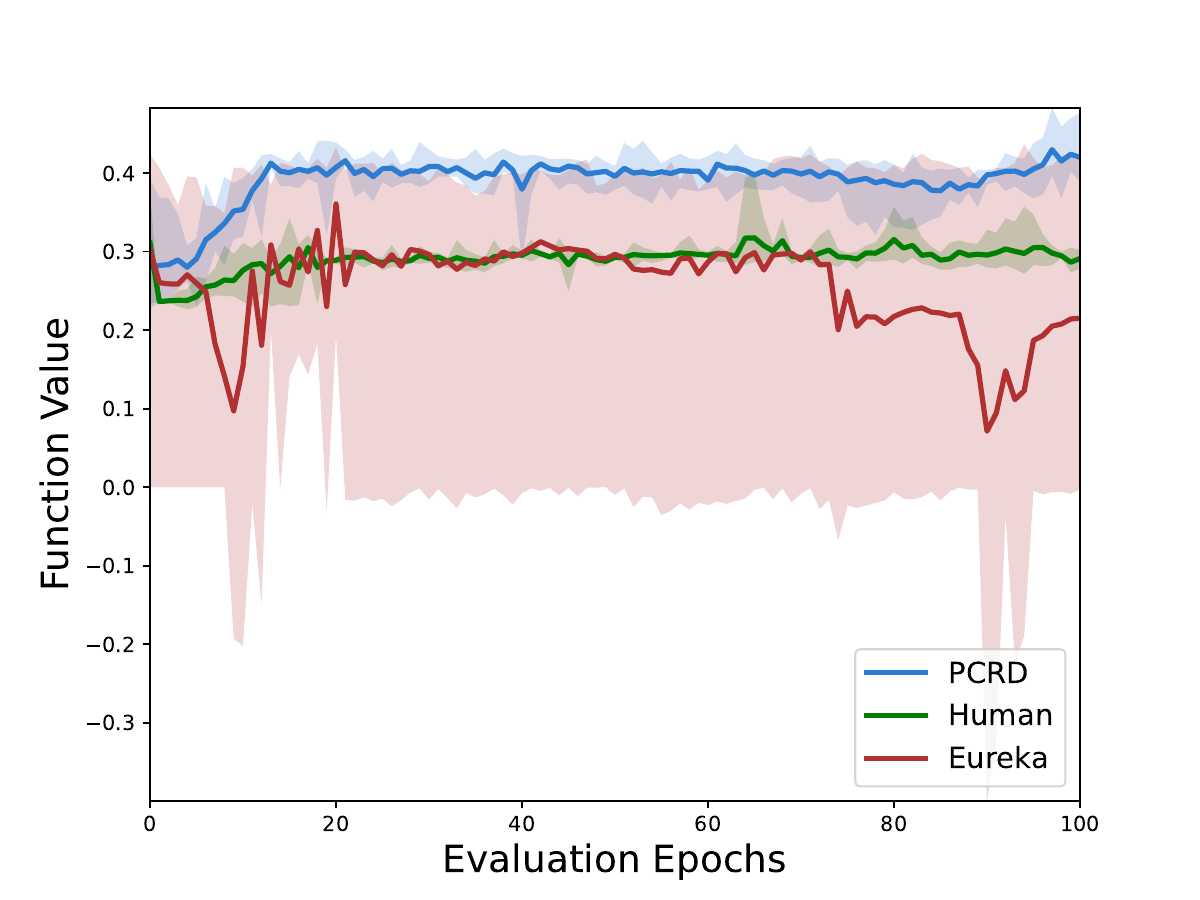}
    \label{figcomparisonf}}    
    \caption{Comparison Experiment in Six Scenarios.}
    \label{figcomparison}
\end{figure*}

\begin{table*}[t]
\centering
\caption{Best Function Value in six scenarios.}
\begin{tabular}{ccccccc}
\toprule
\multirow{2}{*}{Method} & \multicolumn{3}{c}{Single-Object Scenario}                                                 & \multicolumn{3}{c}{Multi-Object Scenario}                                                     \\
\cmidrule(r){2-4} \cmidrule(r){5-7}
                        & Wait                       & Speed             & Mix                       & Wait                     & Speed                    & Mix                      \\
                        \cmidrule(r){1-1} \cmidrule(r){2-4} \cmidrule(r){5-7}
Human                   & 15323.72                   & \textbf{11018.43} & 11115.33                  & 0.6097                   & 0.3795                   & 0.3992                   \\
Eureka                  & 14450.58                   & 10593.54          & 10008.70                  & 0.7150                   & 0.3839                   & 0.4365                   \\
PCRD                    & \textbf{17313.68(11.49\%$\uparrow$)} & 10656.17(-3.39\%$\downarrow$) & \textbf{11904.36(6.63\%$\uparrow$)} & \textbf{0.7155(17.35\%$\uparrow$)} & \textbf{0.4349(12.74\%$\uparrow$)} & \textbf{0.4829(17.33\%$\uparrow$)} \\ \bottomrule
\end{tabular}
\end{table*}

For each objective function, we set the same three coordination methods as scenarios.
\begin{itemize}
    \item [-] \textbf{Wait Scenario:} Trucks can wait at the hub for other trucks. 
    If the trucks depart from the hub at the same time, they will be treated as a platoon until they arrive at the next hub. 
    In this scenario, trucks can plan their waiting time at hubs to form platoons.
    \item [-] \textbf{Speed Scenario:} Trucks can adjust speed on the road to catch or wait for other trucks. 
    If trucks travel at a speed of $v_m$ at the same location (edge or hub), they will be treated as a platoon until the next hub.
    In this scenario, trucks can only adjust their speeds to form platoons.
    \item [-] \textbf{Mix Scenario:} Mix scenario is a combination of the first two scenarios. 
    Trucks can wait at the hub for other trucks or adjust the speed on the road to catch or wait for other trucks. 
    In this scenario, trucks can plan their waiting time and speeds to form platoons.
\end{itemize}
For more details on the scenarios, see previous work \cite{wei2023multi,wei2024multi}. 
Tab. II shows the textual specification $J_{text}$ corresponding to each scenario.
In the single-object scenarios, in order to prevent LLM from overthinking the factors of time delay, we specifically emphasized that time delay should not be considered.

\begin{table}[h]
\centering
\caption{QMIX hyperparameters for platoon coordination.}
\begin{tabular}{cc}
\toprule
Parameter          & Value \\
\cmidrule(r){1-2}
Max Training Steps & 4000  \\
Batch Size         & 24    \\
Replay Buffer Size & 100   \\
Gamma              & 0.99  \\
Learning Rate      & 1e-3  \\
Epsilon Init       & 1     \\
Epsilon Min        & 0.05  \\
Epsilon Steps      & 50000 \\
Evaluation Steps   & 40   \\ \bottomrule
\end{tabular}
\end{table}

\subsubsection{Parameters Setting}
In Algorithm 1, we use GPT-4o as the large language model $\theta_L$ to generate reward function codes and set the maximum iteration $N_{iter}=5$, the quantity of basic generation $k=4$, and the evolutionary quantity $m=1$.
In \eqref{eqmean}, \eqref{eqstd} and \eqref{eqslope}, we set the duration of the early and late stages $\alpha=20, \beta=40$, the volatility threshold $v_{th}=0.5$, and the evaluation times $L=100$.
Similarly to \cite{wei2023multi,wei2024multi}, we used QMIX as the multi-agent deep reinforcement learning algorithm $\Omega_{E_p}(\cdot)$ to train in the platoon coordination environment $E_p$. All experiments used the same hyperparameters for training, which is shown as Tab. IV.

\subsubsection{Baselines}
Our method is compared with the following baseline method, including human design reward and Eureka.
\begin{itemize}
    \item [-] \textbf{Human:} Reward function designed by human experts in \cite{wei2023multi} and \cite{wei2024multi}. 
    Since these reward functions were developed by active reinforcement learning researchers who were also responsible for task design, they reflect expert-level human reward engineering outcomes.
    \item [-] \textbf{Eureka:} Evolution-driven Universal REward Kit for Agent (Eureka) \cite{maeureka} is a previous framework for automatically searching reward functions through LLM. 
    It leverages task descriptions and environment code as direct inputs to generate reward functions, which are then evolved and optimized based on training feedback. 
    Eureka has achieved remarkable success in simple robot control tasks. 
    To adapt it to the complex platoon coordination problem, we tailored Eureka's prompt while preserving its core concept. 
    Modifications involve removing robot task-related content, substituting the fitness function index, and specifying a reward function format compatible with the platoon coordination environment code.
    For fairness, Eureka uses the same search step size $N_{iter}=5$ and search width $k=4$ as PCRD.
\end{itemize}

\begin{figure*}[t]

    \centering
    \subfloat[Single-Object Wait Scenario]{
    \includegraphics[scale=0.28]{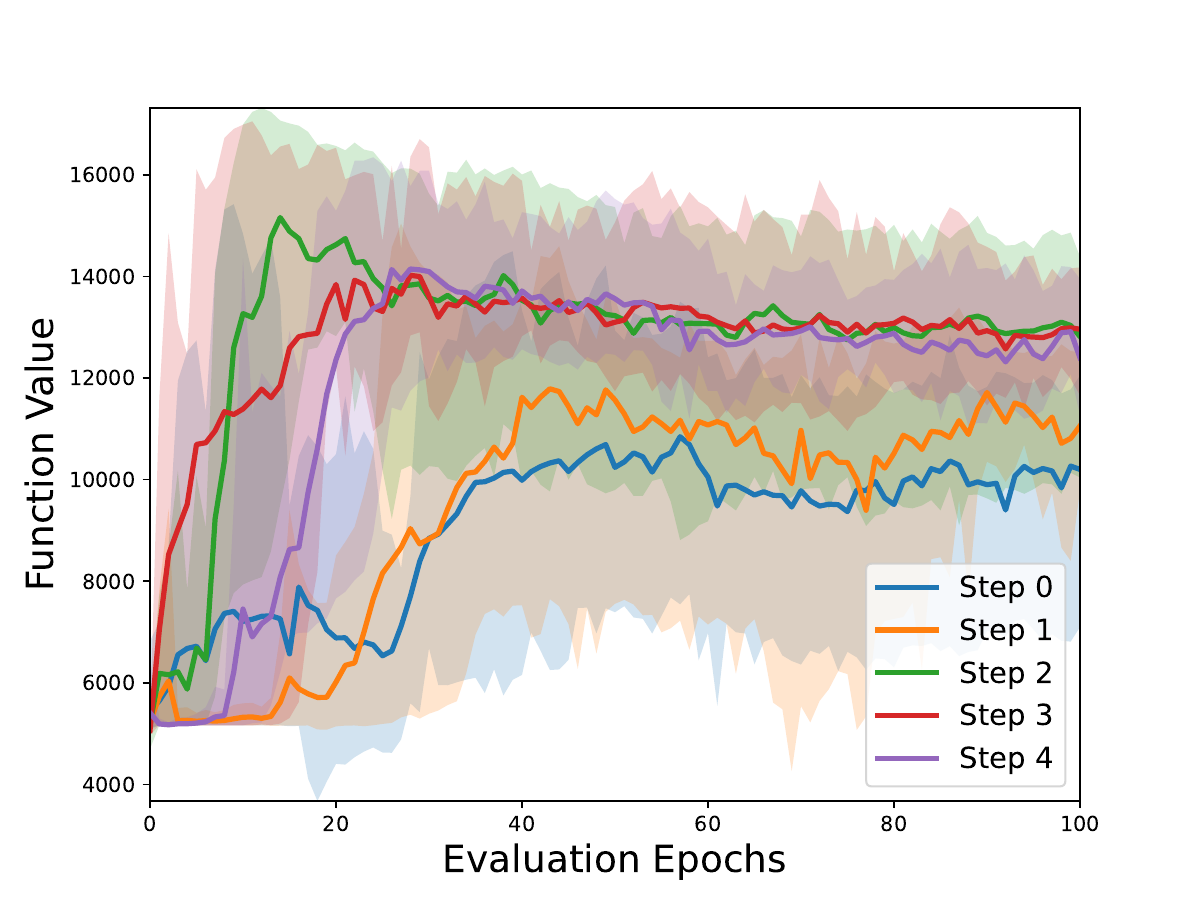}
    \label{figitera}}
    \subfloat[Single-Object Speed Scenario]{
    \includegraphics[scale=0.28]{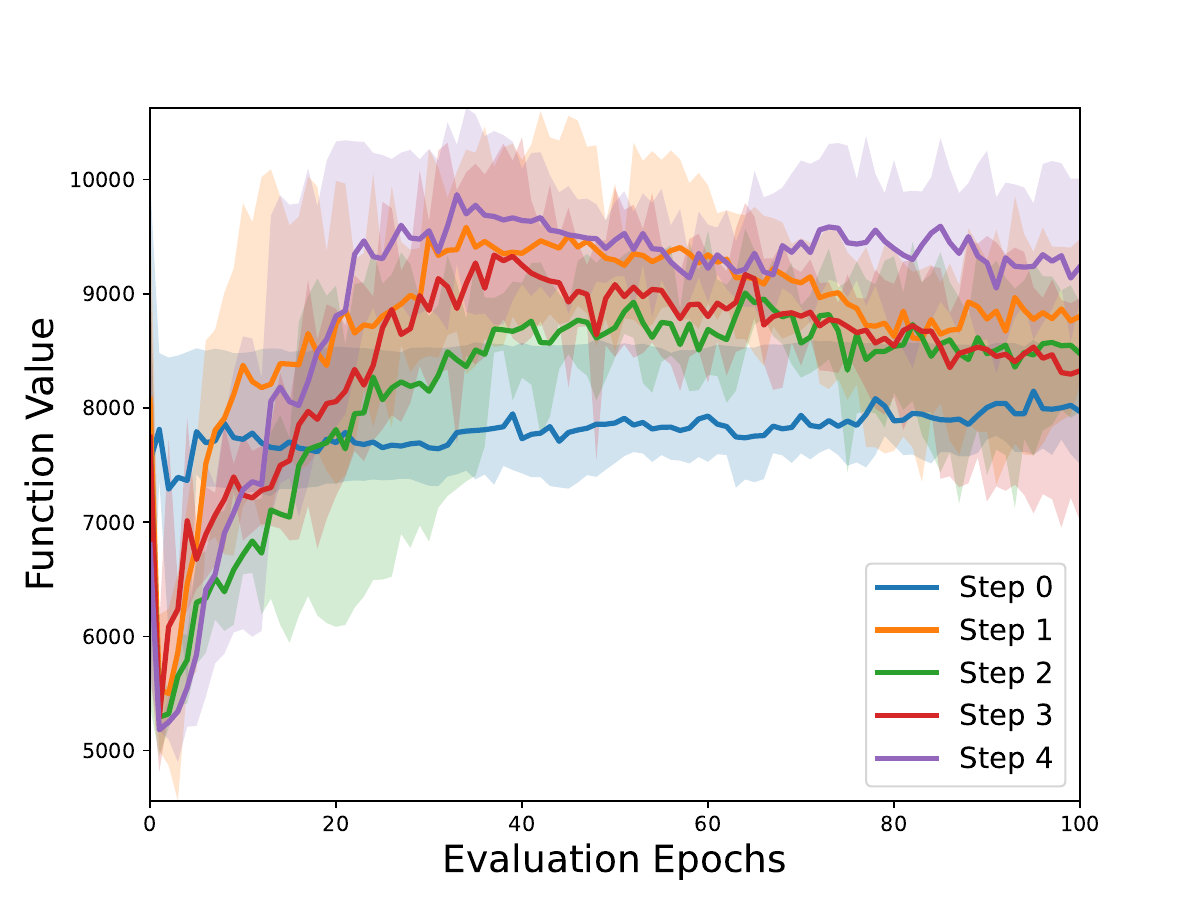}
    \label{figiterb}}
    \subfloat[Single-Object Mix Scenario]{
    \includegraphics[scale=0.28]{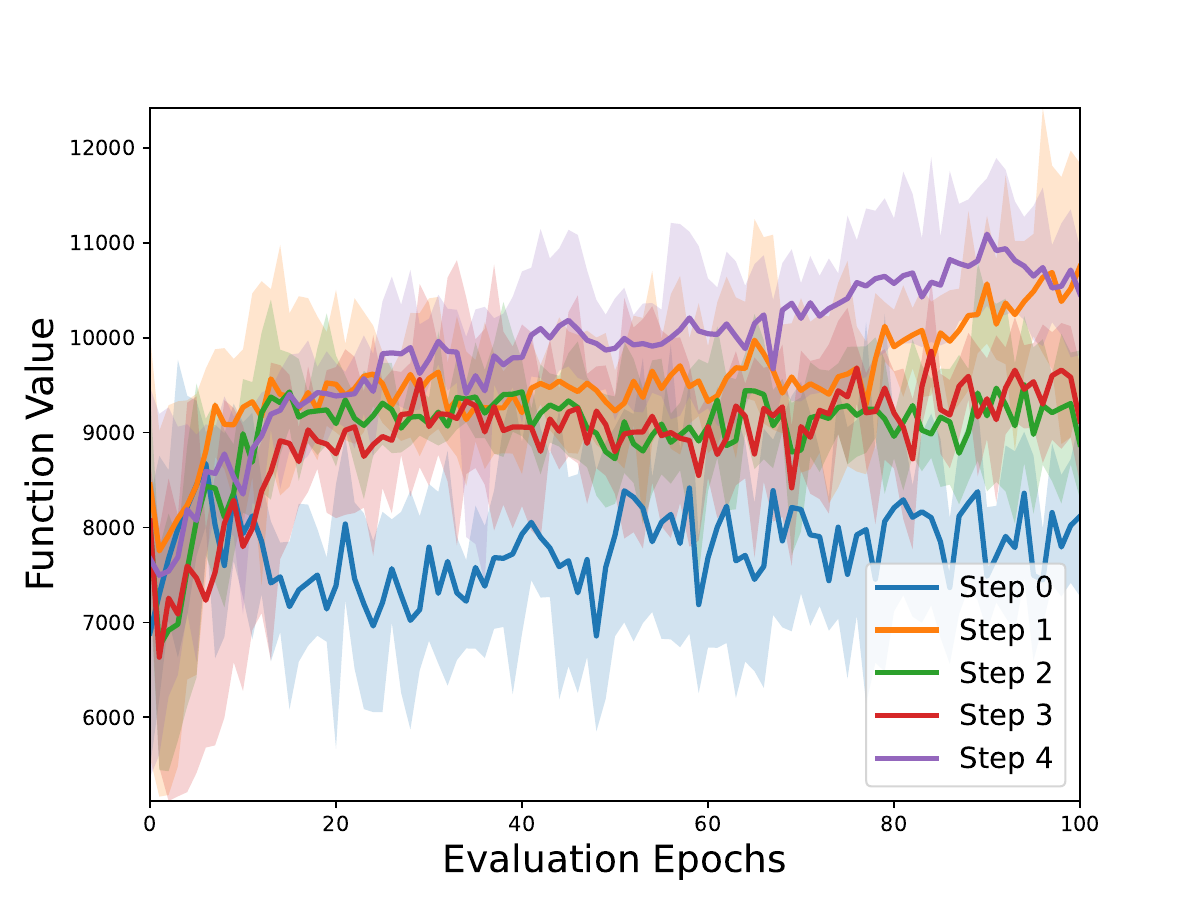}
    \label{figiterc}}\\
    \subfloat[Multi-Object Wait Scenario]{
    \includegraphics[scale=0.28]{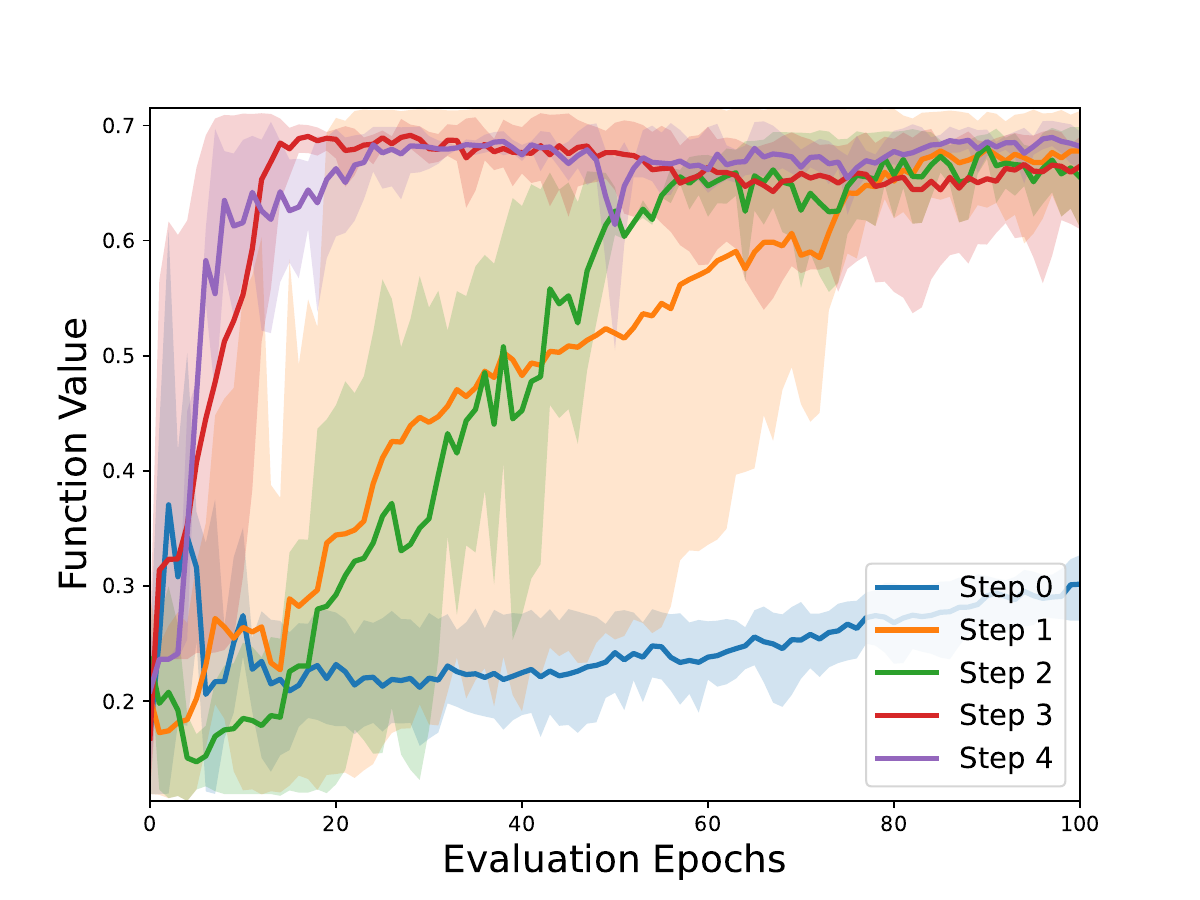}
    \label{figitere}}
    \subfloat[Multi-Object Speed Scenario]{
    \includegraphics[scale=0.28]{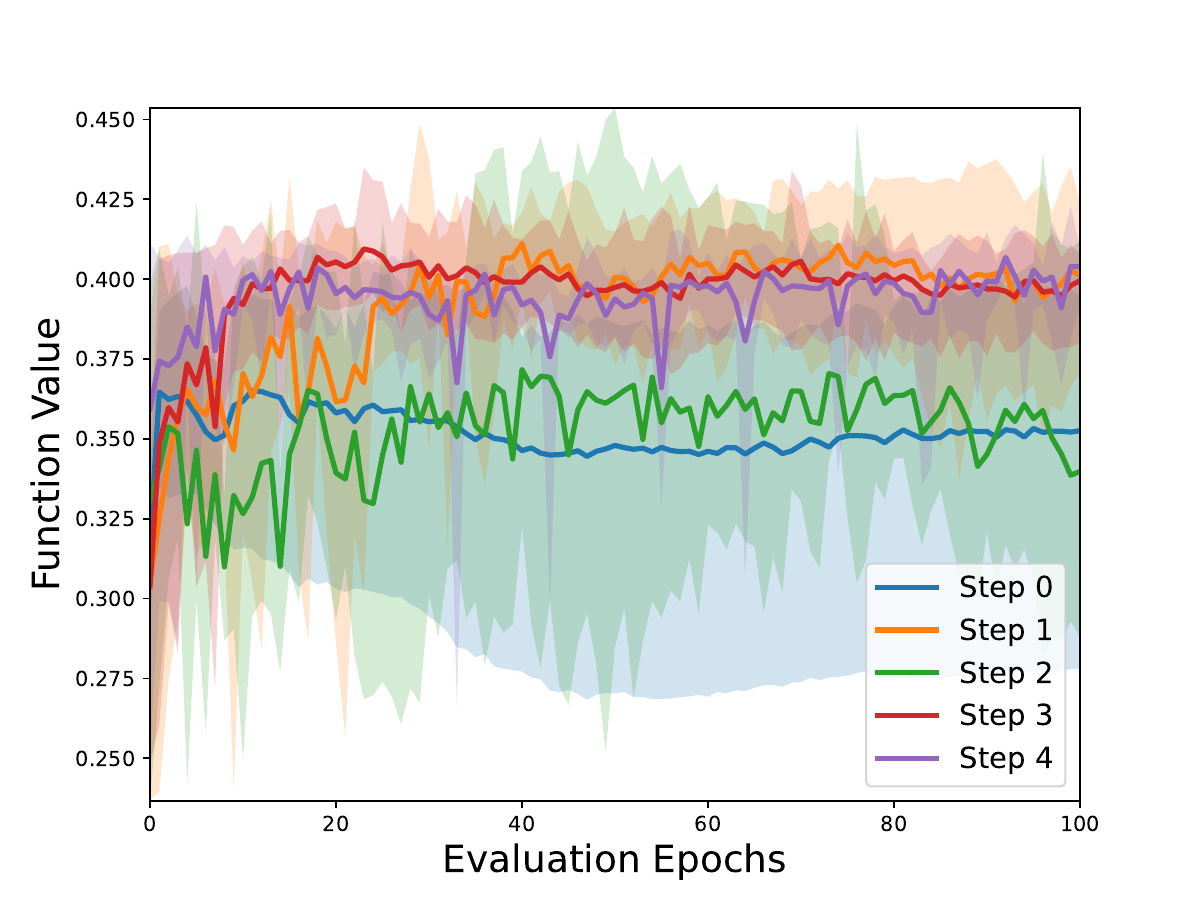}
    \label{figiterd}}
    \subfloat[Multi-Object Mix Scenario]{
    \includegraphics[scale=0.28]{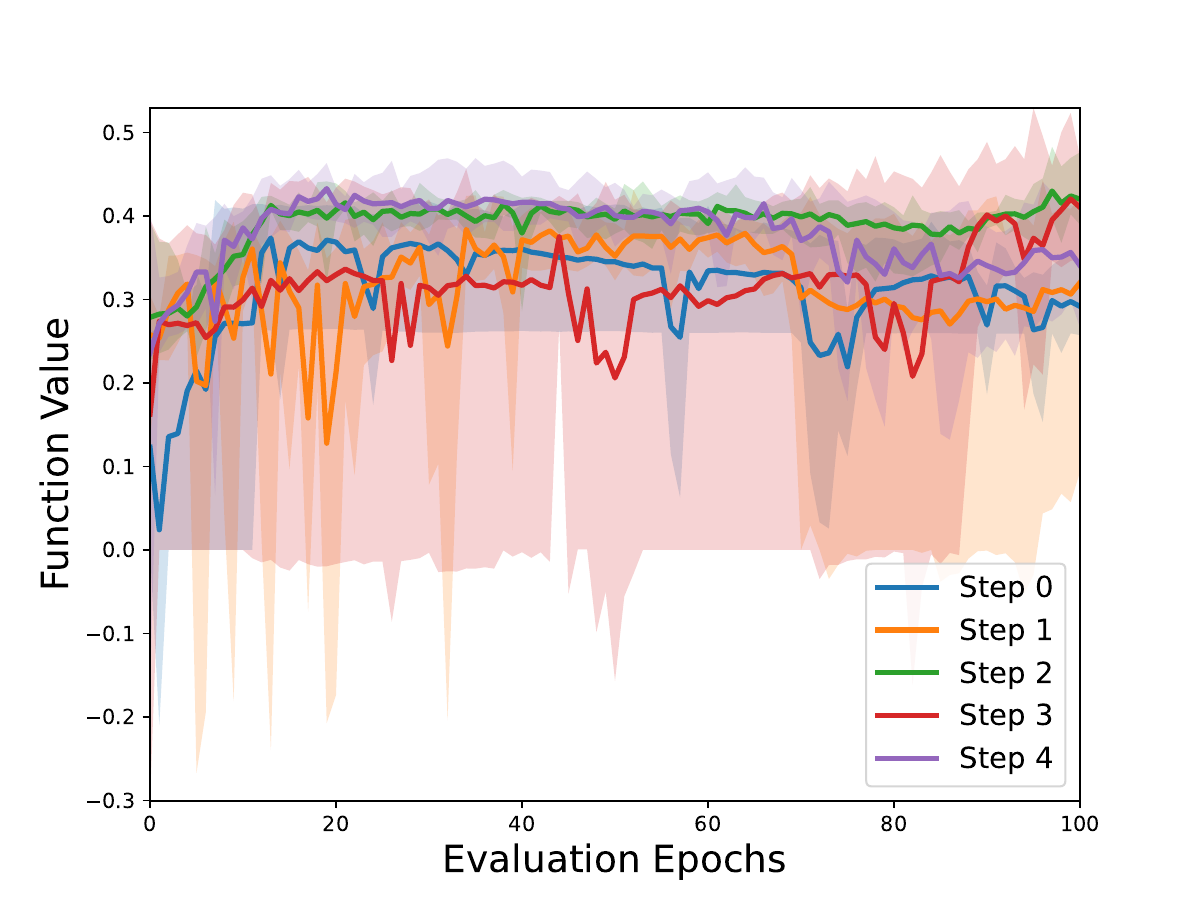}
    \label{figiterf}}
    \caption{Iteration Experiment in Six Scenarios.}
\label{figiter}
\end{figure*}

\subsection{Comparison Experiment}
Under identical freight tasks, scenarios and training parameters, we compared the performance of Human, Eureka, and our method. 
Five experiments were repeated for each reward function. 
The solid line represents the average value of the five experiments, and the shadow represents the upper and lower intervals.
In Fig. \ref{figcomparison}, in the single-object wait and mixed scenarios, PCRD's reward functions surpassed those of human experts. 
In the speed scenario, PCRD's reward functions were on par with human experts.
Eureka's search results are not stable. 
Although the reward function found in the waiting and speed scenarios is roughly the same as that of human experts, it performs poorly in the mix scenario.

In multi-object scenarios, as task difficulty increased, human-designed reward functions did not always perform well. 
The automatic search methods (PCRD and Eureka) outperformed the human experts. 
This is due to the complicity of multi-objective optimization problem (balancing platoon formation and time), where human experts struggled to assign appropriate weights to different objectives. 
In the three multi-object scenarios, PCRD outperformed Eureka, demonstrating its superiority in search efficiency for platoon coordination tasks under the same search steps.
Tab. III presents the objective function values of the three methods under the optimal strategy in each scenario. 
PCRD delivered superior reward functions compared to human experts in five scenarios. 
The reward functions generated by PCRD, through an automated search, achieved an average increase of 10\% in performance compared to those designed by human experts in six scenarios.
These results demonstrate that PCRD, leveraging the AIR and EvoLeap modules, achieves enhanced comprehension of environmental code and task requirements, thereby enabling the effective evolution of superior reward functions.

\begin{table}[t]
\centering
\caption{Evolutionary details in six scenarios.}
\begin{tabular}{ccccccc}
\toprule
\multirow{2}{*}{Iteration} & \multicolumn{3}{c}{Single-Object Scenario} & \multicolumn{3}{c}{Multi-Object Scenario} \\
 \cmidrule(r){2-4} \cmidrule(r){5-7}
                           & Wait    & Speed    & Mix   & Wait    & Speed    & Mix   \\
                           \cmidrule(r){1-1} \cmidrule(r){2-4} \cmidrule(r){5-7}
1                          & L1      & L1       & F2    & L1      & L1       & F3    \\
2                          & L1      & F3       & F2    & F2      & F2       & F3    \\
3                          & F1      & F3       & F2    & F3      & F1       & F2    \\
4                          & F3      & F1       & F3    & F3      & F1       & F3   \\ \bottomrule
\end{tabular}
\end{table}

\begin{figure*}[t]
    \centering
    \includegraphics[scale=0.16]{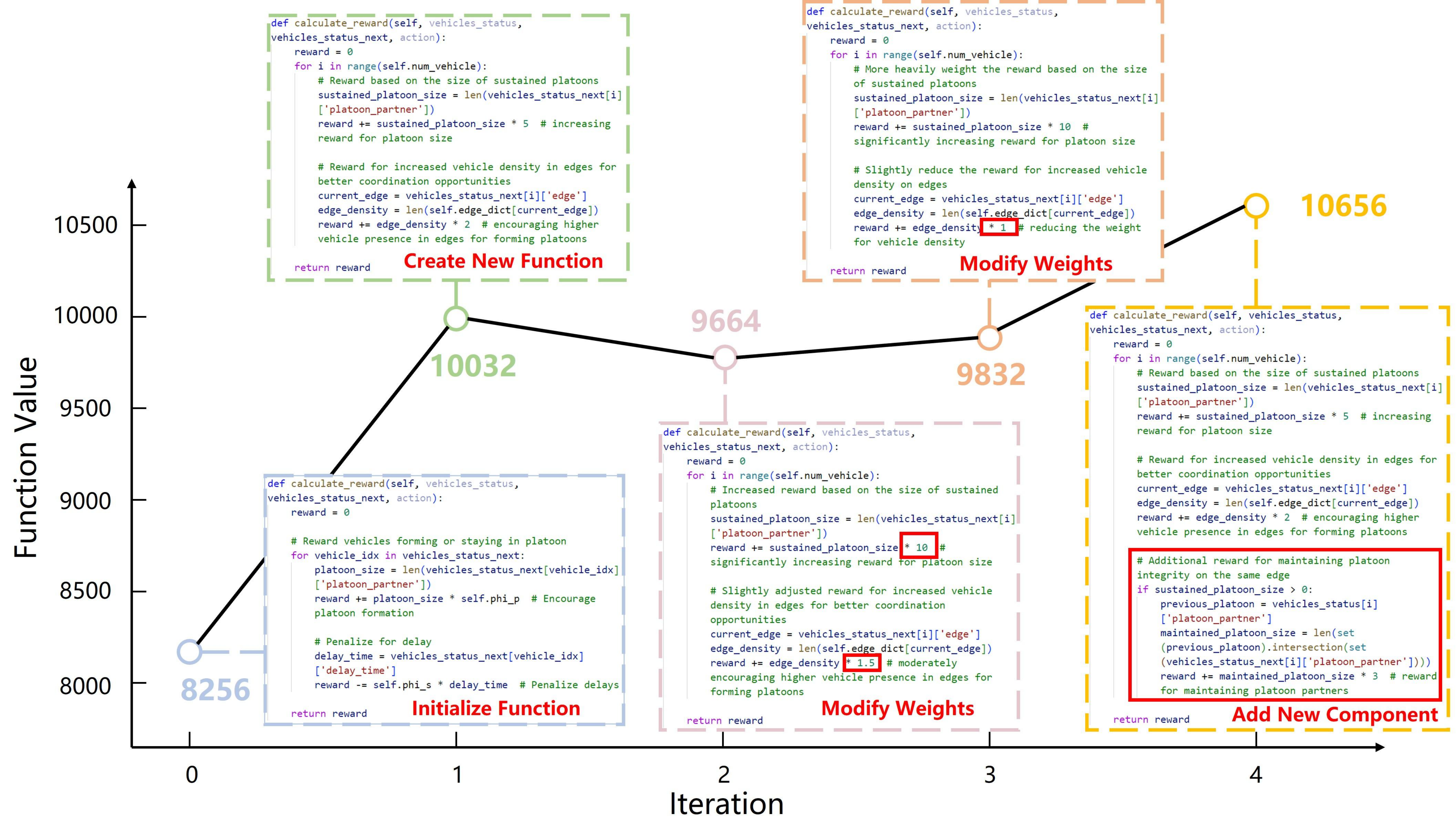}
    \caption{Case study in the single-object speed scenario.}
    \label{case}
\end{figure*}

\subsection{Experiment on Number of Iterations}
To analyze how the number of iterations affects PCRD, we further explore the performance changes of the reward function designed by PCRD during the iteration.
Fig. \ref{figiter} shows the four evolutions in PCRD.
After each evolution, we choose the best-performing reward function for five-round retraining, with the average value plotted as the curve. 
Fig. \ref{figiter} shows that the quality of the reward function gradually improves in the evolution process (step 1, 2) and gradually converges in the late stage of evolution (step 3, 4).
After two rounds of evolution, most scenarios achieved good performance in Step 2. 
PCRD refined the reward function in Steps 3 and 4 for higher performance in single object speed and mix scenarios. 
In multi object scenarios, the post performance of PCRD fluctuates slightly, while the overall region remains stable. 
This indicates that our method can stabilize the performance of evolutionary search and reduce the occurrence of degradation in more complex multi-objective scenarios.

Tab. V shows the evolution method of the reward function that performs best in each round of evolution of PCRD in each scenario. 
In the early stage of evolution (step 1, 2), since the effects of the initially generated reward functions may be of low quality, PCRD tends to search by creating new reward functions. 
In the late stage of evolution (step 3, 4), PCRD used more fine-tuning methods to improve the current reward function, to seek a better breakthrough on the basis of the current reward function.
Therefore, in the CPCP problem, PCRD can search for high-quality reward functions after two iterations. 
And as the number of iterations increases, it can continue to search for better reward functions.

\subsection{Case Study}
As shown in Fig. \ref{case}, to illustrate the details of PCRD's evolution, we present the evolution process of the reward function code in the single-object speed scenario.

\textbf{Iteration 0:} In the initial iteration, LLM generates multiple initial reward functions as search starting points by analyzing code and task requirements.
The first generated reward function is usually not of high quality and needs further iterative evolution. 
As shown in Fig. \ref{case}, the blue box presents the initially generated optimal reward function. 
Although this reward function promotes larger platoon sizes, it still incorporates the redundant time factor in its code.

\textbf{Iteration 1:} The initial reward function template guided the search in the wrong direction. 
So, in the first evolution, PCRD reconstructed a new reward function. 
The green box in Fig. \ref{case} shows this reward function, which encourages a larger platoon size and higher vehicle density on the road.

\textbf{Iteration 2, 3:} During the middle phase of evolutionary exploration, each iteration focused primarily on fine-tuning the weights of the reward function. 
As shown in the pink and orange boxes in Fig. \ref{case}, PCRD adjusted the weights for platoon size and road density. 
However, the performance of these modified reward functions was slightly inferior to that in iteration 1.

\textbf{Iteration 4:} In iteration 4, PCRD enhanced the iteration 1 template by adding a new reward component, thereby improving performance. 
The yellow box in Fig. \ref{case} indicates that PCRD incorporated the reward of maintaining platoon into the iteration 1 reward function.
 
Throughout the entire evolution curve, the evolution process demonstrates convergence and stability. 
This stability arises from our comprehensive exploration strategy, which combines both subtle fine-tuning approaches, such as component simplification and weight modification, and more dramatic exploration techniques, like component addition and complete recreation.

\begin{table*}[t]
\centering
\caption{Code Execution Experiments.}
\begin{tabular}{ccccccccccc}
\toprule
\multirow{2}{*}{Method}         & \multicolumn{3}{c}{Single-Object Scenario} & \multicolumn{3}{c}{Multi-Object Scenario} & \multicolumn{3}{c}{Code Error Type}   & \multirow{2}{*}{Accuracy} \\
                \cmidrule(r){2-4} \cmidrule(r){5-7} \cmidrule(r){8-10}
               & Wait    & Speed    & Mix    & Wait    & Speed    & Mix    & Semantic & Syntax & Running &           \\
               \cmidrule(r){1-1} \cmidrule(r){2-4} \cmidrule(r){5-7} \cmidrule(r){8-10} \cmidrule(r){11-11}
PCRD w/o AIR & 93      & 92       & 91     & 93      & 89       & 87     & 34       & 15      & 6       & 90.83\%  \\
PCRD           & 100     & 99       & 99     & 99      & 99       & 100    & 0        & 4      & 0       & 99.33\%   \\
 \bottomrule
\end{tabular}
\end{table*}

\begin{figure*}[t]
    \centering
    \subfloat[Single-Object Wait Scenario]{
    \includegraphics[scale=0.28]{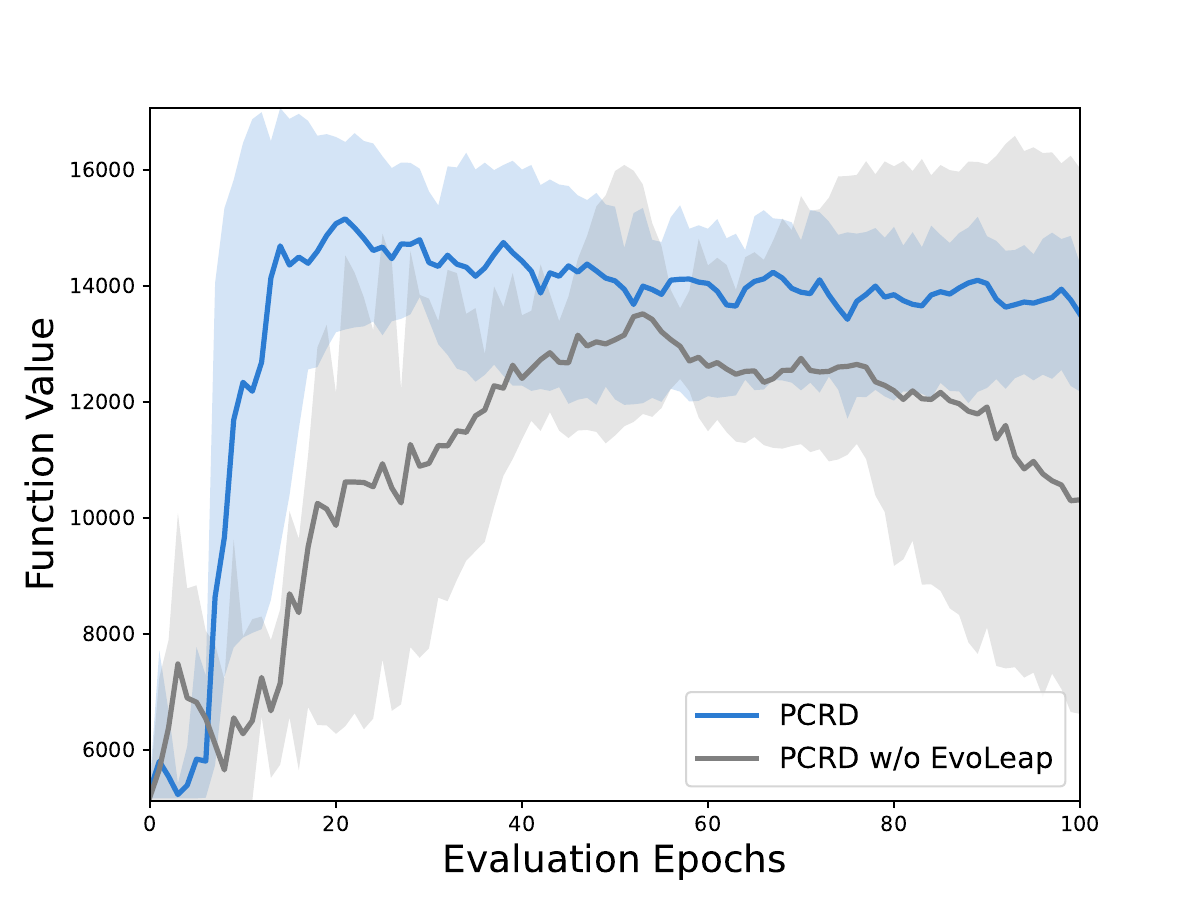}
    \label{figablationa}}
    \subfloat[Single-Object Speed Scenario]{
    \includegraphics[scale=0.28]{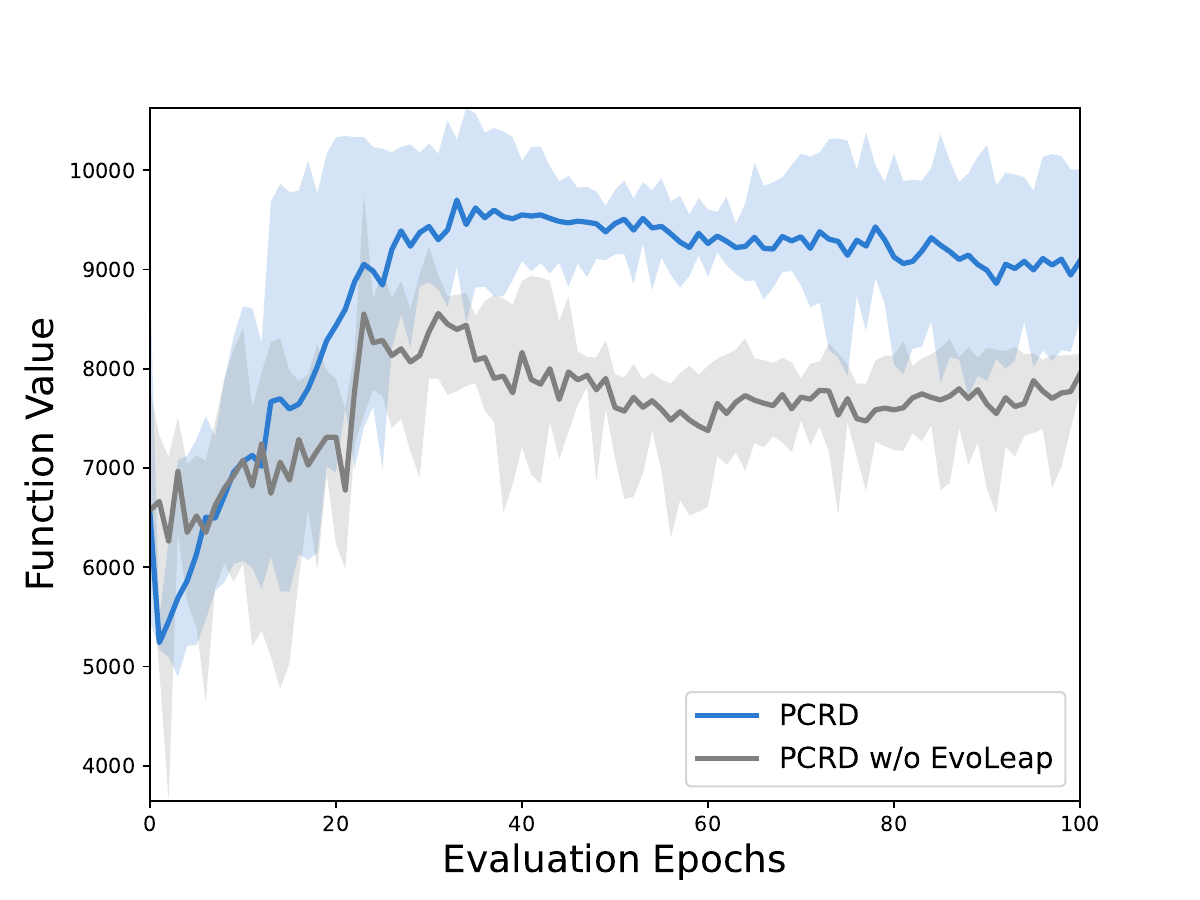}
    \label{figablationb}}
    \subfloat[Single-Object Mix Scenario]{
    \includegraphics[scale=0.28]{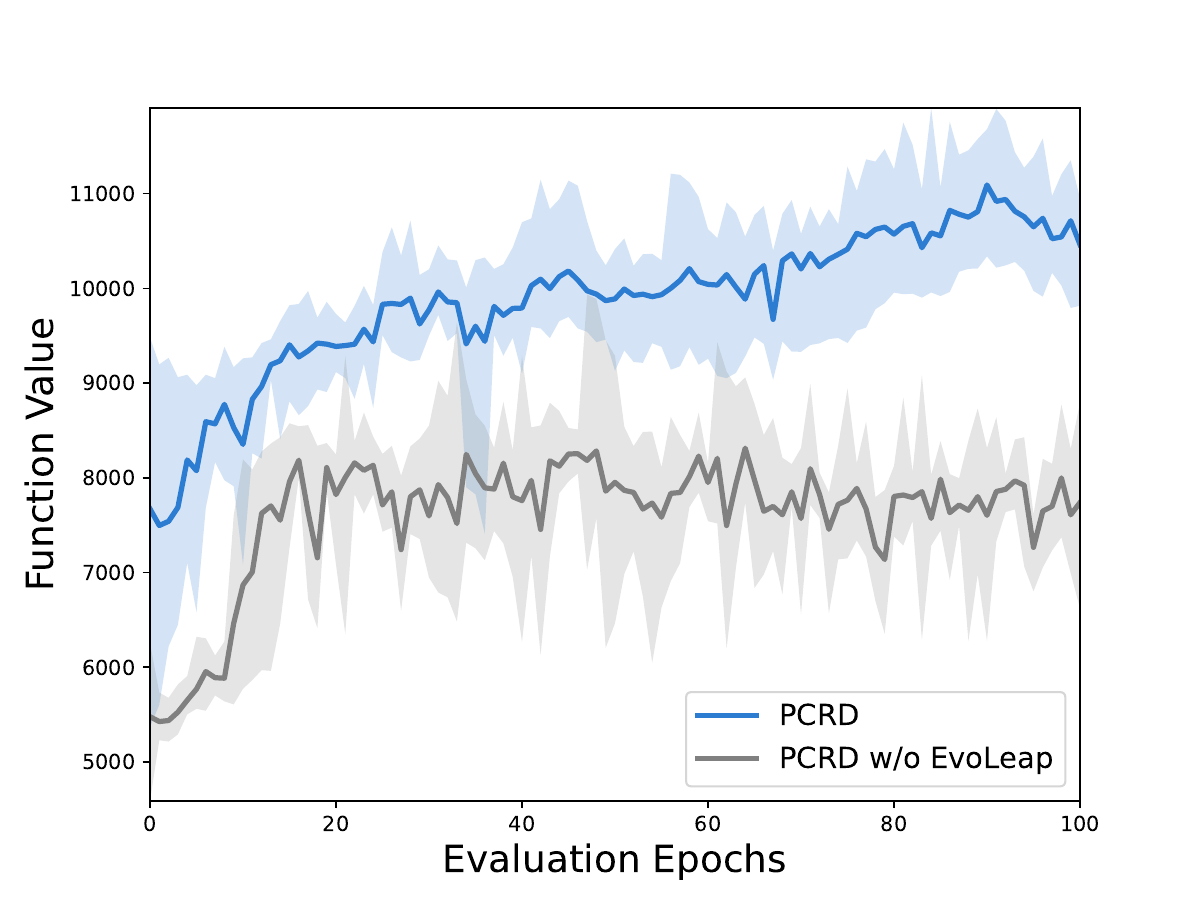}
    \label{figablationc}}\\
    \subfloat[Multi-Object Wait Scenario]{
    \includegraphics[scale=0.28]{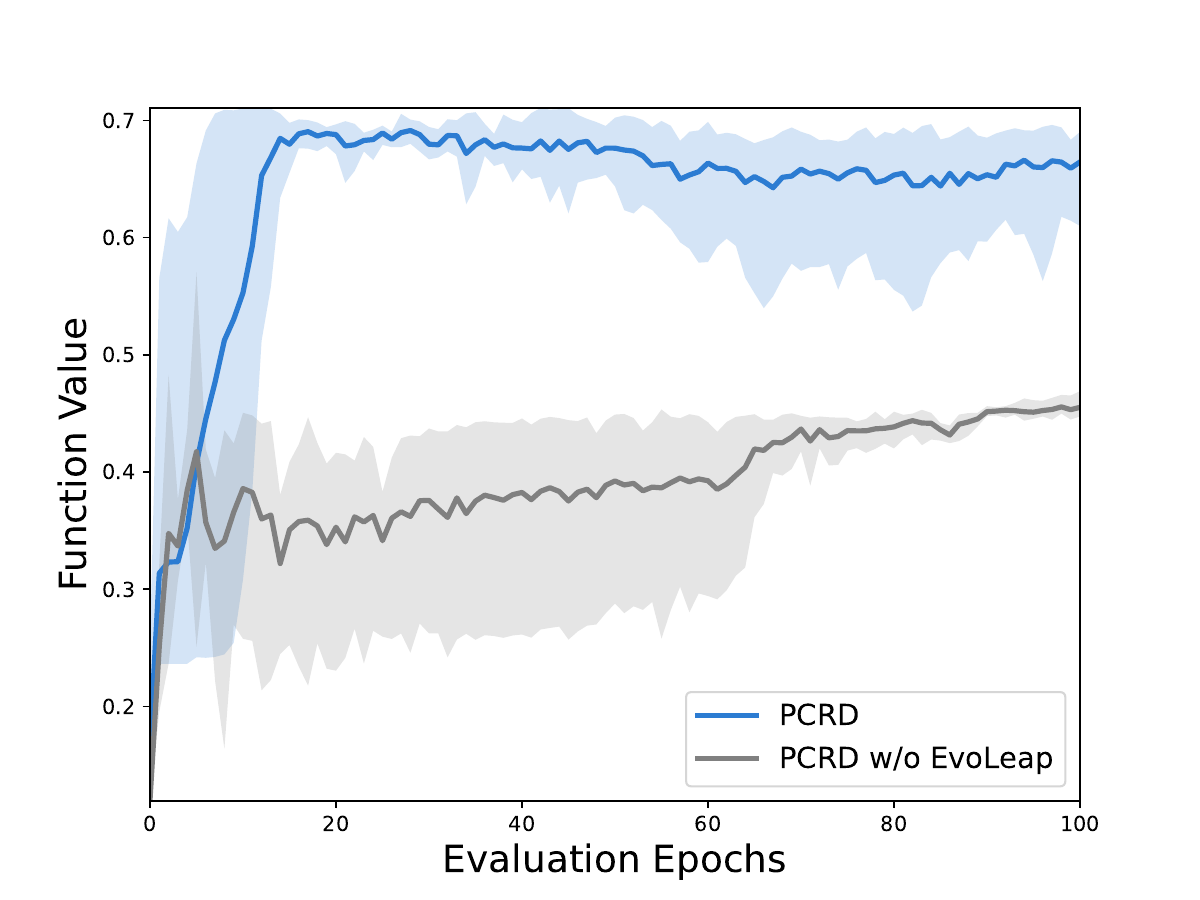}
    \label{figablatione}}
    \subfloat[Multi-Object Speed Scenario]{
    \includegraphics[scale=0.28]{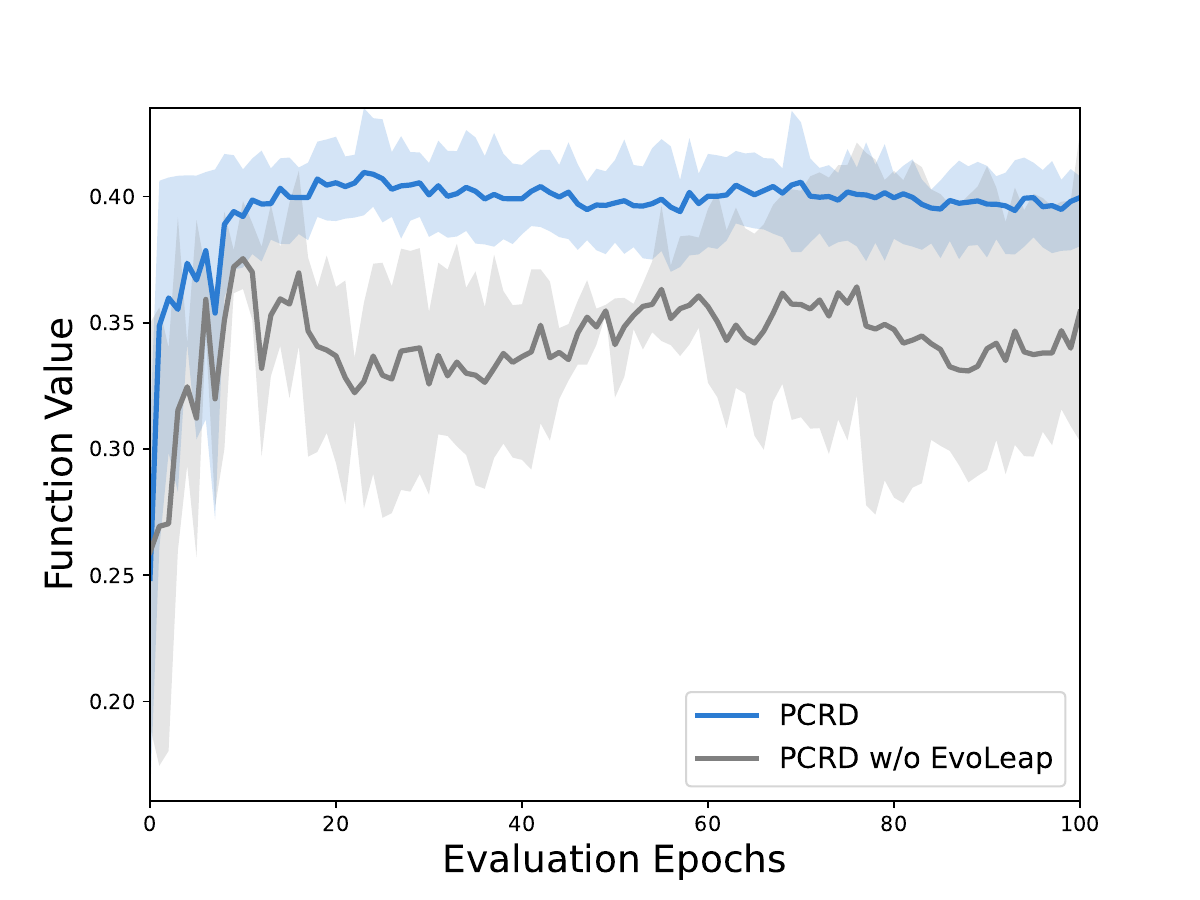}
    \label{figablationd}}
    \subfloat[Multi-Object Mix Scenario]{
    \includegraphics[scale=0.28]{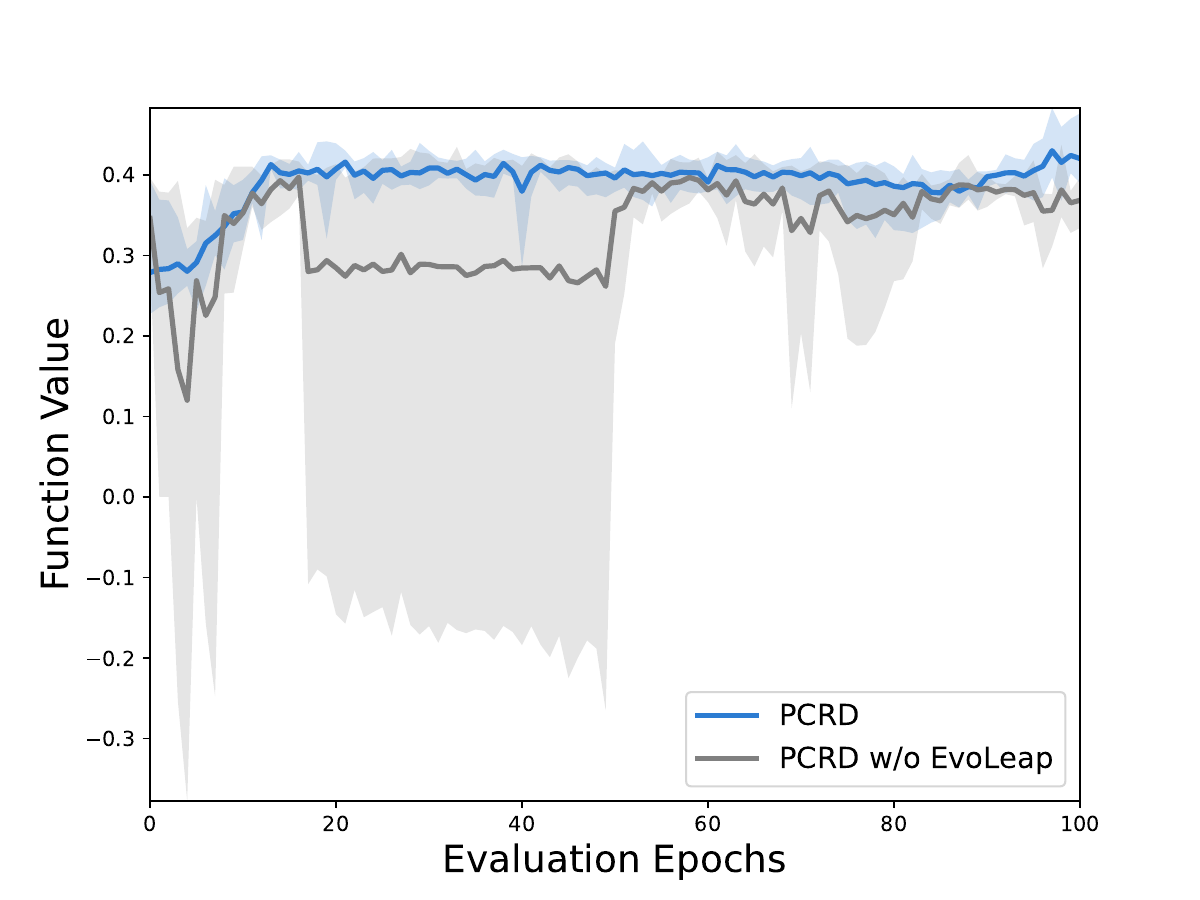}
    \label{figablationf}}
    \caption{Evolution Ablation Experiment in Six Scenarios.}
\label{figablation}
\end{figure*}

\subsection{Ablation Experiments}
In this section, we will discuss the effectiveness of the AIR module and EvoLeap module through ablation experiments.

\subsubsection{Code Execution Error Rate}
As shown in Tab. V, to assess the analysis module's impact on reward function code correctness, we conducted an ablation study. 
We used code and task descriptions from six scenarios as LLM input to determine if the generated reward function code could run properly. 
For each scenario, we conducted 25 experimental repetitions, with LLM generating 4 reward function codes in parallel per experiment. 
This resulted in 100 reward function codes per scenario, totaling 600 cases.

Tab. VI presents the code results for PCRD with and without the analysis module. Without the analysis module, PCRD achieved almost 90\% accuracy in all scenarios. 
With the analysis module, the accuracy reached almost 100\% in all scenarios. 
We also categorized and counted code errors, identifying three primary types: semantic errors, syntax errors, and runtime errors.
\begin{itemize}
    \item \textbf{Semantic Errors:} These primarily involved the use of undefined or nonexistent variables.
    \item \textbf{Syntax Errors:} These mainly consisted of deviations from basic syntax rules, such as missing parentheses or brackets.
    \item \textbf{Runtime Errors:} These primarily involved improper division operations, such as division by zero during execution.
\end{itemize}
We found that the analysis module significantly enhanced code accuracy by substantially reducing semantic and runtime errors. This indicates that analyzing code through the analysis module before generating reward functions effectively improves LLM's code understanding and minimizes errors related to undefined variables or logical issues. 
After adding the analysis module, a few syntax errors still existed in the code, mainly due to LLM sometimes omitting ')' in its output. 
This issue is usually related to the LLM's performance.

\subsubsection{Evolution Performance}
Fig. \ref{figablation} illustrates how EvoLeap affects the search performance of the reward function. 
We replaced EvoLeap with Eureka's evolutionary prompt. 
To ensure fairness, both methods started from the same initial reward functions, with 4 evolutions and a search breadth of 4.
The optimal reward functions found by both methods were then selected in six scenarios. 
Fig. \ref{figablation} shows that PCRD with EvoLeap performs better. Without EvoLeap, the PCRD search is less stable. 
Although it can find reward functions comparable to PCRD in Fig. \ref{figablationa} and Fig. \ref{figablationf}, its performance is relatively poor in other scenarios.

\section{Conclusion}
In this paper, we introduce a platoon coordination reward design problem in the context of the cooperative platoon coordination problem.
Then we propose the PCRD framework, an LLM-based reward design framework for training RL policies to address cooperative platoon coordination problems. 
PCRD comprises three essential elements: the AIR module, MADRL parallel training, and the EvoLeap module. 
The AIR module utilizes the chain of thought to analyze the environment code across multiple dimensions and generates initial reward function codes based on task requirements. 
These codes are stored in a reward function code pool and used as starting points for reinforcement learning training. 
The system then conducts parallel training with these codes and the results are filtered and sorted to select the optimal reward function as the evolutionary template. 
The EvoLeap module evolves this template directionally based on training feedback, fine-tuning and refactoring the reward function to balance search breadth and stability. 
Through multiple evolution cycles, PCRD continuously searches for improved reward functions.
Experiments on the Yangtze River Delta region simulation have shown that the PCRD framework represents a significant advance in automating the design of reward functions for platoon coordination problems. 
It shows superior performance compared to expert-designed reward functions and has the potential to streamline the development of RL policies for complex coordination tasks. 
Future research will focus on further refining the PCRD framework to enhance its search efficiency and explore its application to more intricate and diverse platoon coordination scenarios.

\bibliographystyle{IEEEtran}
\small
\bibliography{main}

\begin{thebibliography}{10}
\providecommand{\url}[1]{#1}
\csname url@samestyle\endcsname
\providecommand{\newblock}{\relax}
\providecommand{\bibinfo}[2]{#2}
\providecommand{\BIBentrySTDinterwordspacing}{\spaceskip=0pt\relax}
\providecommand{\BIBentryALTinterwordstretchfactor}{4}
\providecommand{\BIBentryALTinterwordspacing}{\spaceskip=\fontdimen2\font plus
\BIBentryALTinterwordstretchfactor\fontdimen3\font minus \fontdimen4\font\relax}
\providecommand{\BIBforeignlanguage}[2]{{%
\expandafter\ifx\csname l@#1\endcsname\relax
\typeout{** WARNING: IEEEtran.bst: No hyphenation pattern has been}%
\typeout{** loaded for the language `#1'. Using the pattern for}%
\typeout{** the default language instead.}%
\else
\language=\csname l@#1\endcsname
\fi
#2}}
\providecommand{\BIBdecl}{\relax}
\BIBdecl

\bibitem{jiang2024cooperative}
L.~Jiang, J.~Kheyrollahi, C.~R. Koch, and M.~Shahbakhti, ``Cooperative truck platooning trial on canadian public highway under commercial operation in winter driving conditions,'' \emph{Proceedings of the Institution of Mechanical Engineers, Part D: Journal of Automobile Engineering}, p. 09544070241245477, 2024.

\bibitem{tsugawa2014results}
S.~Tsugawa, ``Results and issues of an automated truck platoon within the energy its project,'' in \emph{2014 IEEE Intelligent Vehicles Symposium Proceedings}.\hskip 1em plus 0.5em minus 0.4em\relax IEEE, 2014, pp. 642--647.

\bibitem{van2017fuel}
S.~Van De~Hoef, K.~H. Johansson, and D.~V. Dimarogonas, ``Fuel-efficient en route formation of truck platoons,'' \emph{IEEE Transactions on Intelligent Transportation Systems}, vol.~19, no.~1, pp. 102--112, 2017.

\bibitem{xu2022optimizing}
W.~Xu, T.~Cui, and M.~Chen, ``Optimizing two-truck platooning with deadlines,'' \emph{IEEE Transactions on Intelligent Transportation Systems}, vol.~24, no.~1, pp. 694--705, 2022.

\bibitem{hu2024optimal}
Q.~Hu, W.~Gu, L.~Wu, and L.~Zhang, ``Optimal autonomous truck platooning with detours, nonlinear costs, and a platoon size constraint,'' \emph{Transportation Research Part E: Logistics and Transportation Review}, vol. 186, p. 103545, 2024.

\bibitem{hoef2019predictive}
S.~V.~D. Hoef, J.~M{\aa}rtensson, D.~V. Dimarogonas, and K.~H. Johansson, ``A predictive framework for dynamic heavy-duty vehicle platoon coordination,'' \emph{ACM Transactions on Cyber-Physical Systems}, vol.~4, no.~1, pp. 1--25, 2019.

\bibitem{liang2015heavy}
K.-Y. Liang, J.~M{\aa}rtensson, and K.~H. Johansson, ``Heavy-duty vehicle platoon formation for fuel efficiency,'' \emph{IEEE Transactions on Intelligent Transportation Systems}, vol.~17, no.~4, pp. 1051--1061, 2015.

\bibitem{choi2024optimizing}
J.~Choi and B.~Do~Chung, ``Optimizing vehicle route, schedule, and platoon formation considering time-dependent traffic congestion,'' \emph{Computers \& Industrial Engineering}, vol. 192, p. 110205, 2024.

\bibitem{bouchery2022coalition}
Y.~Bouchery, B.~Hezarkhani, and G.~Stauffer, ``Coalition formation and cost sharing for truck platooning,'' \emph{Transportation Research Part B: Methodological}, vol. 165, pp. 15--34, 2022.

\bibitem{bai2023large}
T.~Bai, A.~Johansson, K.~H. Johansson, and J.~M{\aa}rtensson, ``Large-scale multi-fleet platoon coordination: A dynamic programming approach,'' \emph{IEEE Transactions on Intelligent Transportation Systems}, 2023.

\bibitem{larsson2015vehicle}
E.~Larsson, G.~Sennton, and J.~Larson, ``The vehicle platooning problem: Computational complexity and heuristics,'' \emph{Transportation Research Part C: Emerging Technologies}, vol.~60, pp. 258--277, 2015.

\bibitem{van2017efficient}
S.~Van~de Hoef, K.~H. Johansson, and D.~V. Dimarogonas, ``Efficient dynamic programming solution to a platoon coordination merge problem with stochastic travel times,'' \emph{IFAC-PapersOnLine}, vol.~50, no.~1, pp. 4228--4233, 2017.

\bibitem{xiong2024approximate}
X.~Xiong, M.~Wang, D.~Sun, and L.~Jin, ``An approximate dynamic programming approach to vehicle platooning coordination in networks,'' \emph{IEEE Transactions on Intelligent Transportation Systems}, 2024.

\bibitem{bai2021event}
T.~Bai, A.~Johansson, K.~H. Johansson, and J.~M{\aa}rtensson, ``Event-triggered distributed model predictive control for platoon coordination at hubs in a transport system,'' in \emph{2021 60th IEEE Conference on Decision and Control (CDC)}.\hskip 1em plus 0.5em minus 0.4em\relax IEEE, 2021, pp. 1198--1204.

\bibitem{pan2024towards}
Y.~Pan, J.~Lei, P.~Yi, L.~Guo, and H.~Chen, ``Towards cooperative driving among heterogeneous cavs: A safe multi-agent reinforcement learning approach,'' \emph{IEEE Transactions on Intelligent Vehicles}, 2024.

\bibitem{kumaravel2021optimal}
S.~D. Kumaravel, A.~A. Malikopoulos, and R.~Ayyagari, ``Optimal coordination of platoons of connected and automated vehicles at signal-free intersections,'' \emph{IEEE Transactions on Intelligent Vehicles}, vol.~7, no.~2, pp. 186--197, 2021.

\bibitem{pan2024heterogeneous}
Y.~Pan, J.~Lei, and P.~Yi, ``Heterogeneous multi-agent reinforcement learning based on adaptive curiosity for traffic signal control,'' in \emph{2024 American Control Conference (ACC)}.\hskip 1em plus 0.5em minus 0.4em\relax IEEE, 2024, pp. 239--244.

\bibitem{wei2023multi}
D.~Wei, P.~Yi, and J.~Lei, ``Multi-agent deep reinforcement learning for large-scale platoon coordination with partial information at hubs,'' in \emph{2023 62nd IEEE Conference on Decision and Control (CDC)}.\hskip 1em plus 0.5em minus 0.4em\relax IEEE, 2023, pp. 6242--6248.

\bibitem{singh2009rewards}
S.~Singh, R.~L. Lewis, and A.~G. Barto, ``Where do rewards come from,'' in \emph{Proceedings of the annual conference of the cognitive science society}.\hskip 1em plus 0.5em minus 0.4em\relax Cognitive Science Society, 2009, pp. 2601--2606.

\bibitem{laud2004theory}
A.~D. Laud, \emph{Theory and application of reward shaping in reinforcement learning}.\hskip 1em plus 0.5em minus 0.4em\relax University of Illinois at Urbana-Champaign, 2004.

\bibitem{johansson2023hub}
A.~Johansson, E.~Nekouei, X.~Sun, K.~H. Johansson, and J.~M{\aa}rtensson, ``Hub-based platoon formation: Optimal release policies and approximate solutions,'' \emph{IEEE Transactions on Intelligent Transportation Systems}, vol.~25, no.~6, pp. 5755--5766, 2023.

\bibitem{booth2023perils}
S.~Booth, W.~B. Knox, J.~Shah, S.~Niekum, P.~Stone, and A.~Allievi, ``The perils of trial-and-error reward design: misdesign through overfitting and invalid task specifications,'' in \emph{Proceedings of the AAAI Conference on Artificial Intelligence}, vol.~37, no.~5, 2023, pp. 5920--5929.

\bibitem{yu2023language}
W.~Yu, N.~Gileadi, C.~Fu, S.~Kirmani, K.-H. Lee, M.~G. Arenas, H.-T.~L. Chiang, T.~Erez, L.~Hasenclever, J.~Humplik \emph{et~al.}, ``Language to rewards for robotic skill synthesis,'' in \emph{Conference on Robot Learning}.\hskip 1em plus 0.5em minus 0.4em\relax PMLR, 2023, pp. 374--404.

\bibitem{maeureka}
Y.~J. Ma, W.~Liang, G.~Wang, D.-A. Huang, O.~Bastani, D.~Jayaraman, Y.~Zhu, L.~Fan, and A.~Anandkumar, ``Eureka: Human-level reward design via coding large language models,'' in \emph{The Twelfth International Conference on Learning Representations}.

\bibitem{li2024auto}
H.~Li, X.~Yang, Z.~Wang, X.~Zhu, J.~Zhou, Y.~Qiao, X.~Wang, H.~Li, L.~Lu, and J.~Dai, ``Auto mc-reward: Automated dense reward design with large language models for minecraft,'' in \emph{Proceedings of the IEEE/CVF Conference on Computer Vision and Pattern Recognition}, 2024, pp. 16\,426--16\,435.

\bibitem{xietext2reward}
T.~Xie, S.~Zhao, C.~H. Wu, Y.~Liu, Q.~Luo, V.~Zhong, Y.~Yang, and T.~Yu, ``Text2reward: Reward shaping with language models for reinforcement learning,'' in \emph{The Twelfth International Conference on Learning Representations}.

\bibitem{han2024autoreward}
X.~Han, Q.~Yang, X.~Chen, Z.~Cai, X.~Chu, and M.~Zhu, ``Autoreward: Closed-loop reward design with large language models for autonomous driving,'' \emph{IEEE Transactions on Intelligent Vehicles}, 2024.

\bibitem{ji2023towards}
Z.~Ji, T.~Yu, Y.~Xu, N.~Lee, E.~Ishii, and P.~Fung, ``Towards mitigating llm hallucination via self reflection,'' in \emph{Findings of the Association for Computational Linguistics: EMNLP 2023}, 2023, pp. 1827--1843.

\bibitem{liu2024exploring}
F.~Liu, Y.~Liu, L.~Shi, H.~Huang, R.~Wang, Z.~Yang, L.~Zhang, Z.~Li, and Y.~Ma, ``Exploring and evaluating hallucinations in llm-powered code generation,'' \emph{arXiv preprint arXiv:2404.00971}, 2024.

\bibitem{yao2023llm}
J.-Y. Yao, K.-P. Ning, Z.-H. Liu, M.-N. Ning, Y.-Y. Liu, and L.~Yuan, ``Llm lies: Hallucinations are not bugs, but features as adversarial examples,'' \emph{arXiv preprint arXiv:2310.01469}, 2023.

\bibitem{sun2024large}
S.~Sun, R.~Liu, J.~Lyu, J.-W. Yang, L.~Zhang, and X.~Li, ``A large language model-driven reward design framework via dynamic feedback for reinforcement learning,'' \emph{arXiv preprint arXiv:2410.14660}, 2024.

\bibitem{lesch2021overview}
V.~Lesch, M.~Breitbach, M.~Segata, C.~Becker, S.~Kounev, and C.~Krupitzer, ``An overview on approaches for coordination of platoons,'' \emph{IEEE Transactions on Intelligent Transportation Systems}, vol.~23, no.~8, pp. 10\,049--10\,065, 2021.

\bibitem{larson2013coordinated}
J.~Larson, C.~Kammer, K.-Y. Liang, and K.~H. Johansson, ``Coordinated route optimization for heavy-duty vehicle platoons,'' in \emph{16th International IEEE Conference on Intelligent Transportation Systems (ITSC 2013)}.\hskip 1em plus 0.5em minus 0.4em\relax IEEE, 2013, pp. 1196--1202.

\bibitem{liang2013fuel}
K.-Y. Liang, J.~M{\aa}rtensson, and K.~H. Johansson, ``When is it fuel efficient for a heavy duty vehicle to catch up with a platoon?'' \emph{IFAC Proceedings Volumes}, vol.~46, no.~21, pp. 738--743, 2013.

\bibitem{van2015fuel}
S.~Van De~Hoef, K.~H. Johansson, and D.~V. Dimarogonas, ``Fuel-optimal centralized coordination of truck platooning based on shortest paths,'' in \emph{2015 american control conference (acc)}.\hskip 1em plus 0.5em minus 0.4em\relax IEEE, 2015, pp. 3740--3745.

\bibitem{zhang2017freight}
W.~Zhang, E.~Jenelius, and X.~Ma, ``Freight transport platoon coordination and departure time scheduling under travel time uncertainty,'' \emph{Transportation Research Part E: Logistics and Transportation Review}, vol.~98, pp. 1--23, 2017.

\bibitem{johansson2018multi}
A.~Johansson, E.~Nekouei, K.~H. Johansson, and J.~M{\aa}rtensson, ``Multi-fleet platoon matching: A game-theoretic approach,'' in \emph{2018 21st international conference on intelligent transportation systems (itsc)}.\hskip 1em plus 0.5em minus 0.4em\relax IEEE, 2018, pp. 2980--2985.

\bibitem{boysen2018identical}
N.~Boysen, D.~Briskorn, and S.~Schwerdfeger, ``The identical-path truck platooning problem,'' \emph{Transportation Research Part B: Methodological}, vol. 109, pp. 26--39, 2018.

\bibitem{johansson2021strategic}
A.~Johansson, E.~Nekouei, K.~H. Johansson, and J.~M{\aa}rtensson, ``Strategic hub-based platoon coordination under uncertain travel times,'' \emph{IEEE Transactions on Intelligent Transportation Systems}, vol.~23, no.~7, pp. 8277--8287, 2021.

\bibitem{johansson2022platoon}
A.~Johansson, T.~Bai, K.~H. Johansson, and J.~M{\aa}rtensson, ``Platoon cooperation across carriers: From system architecture to coordination,'' \emph{IEEE Intelligent Transportation Systems Magazine}, vol.~15, no.~3, pp. 132--144, 2022.

\bibitem{xu2022truck}
M.~Xu, X.~Yan, and Y.~Yin, ``Truck routing and platooning optimization considering drivers’ mandatory breaks,'' \emph{Transportation Research Part C: Emerging Technologies}, vol. 143, p. 103809, 2022.

\bibitem{zhang2017platoon}
W.~Zhang, M.~Sundberg, and A.~Karlstr{\"o}m, ``Platoon coordination with time windows: an operational perspective,'' \emph{Transportation Research Procedia}, vol.~27, pp. 357--364, 2017.

\bibitem{chu2019multi}
T.~Chu, J.~Wang, L.~Codec{\`a}, and Z.~Li, ``Multi-agent deep reinforcement learning for large-scale traffic signal control,'' \emph{IEEE transactions on intelligent transportation systems}, vol.~21, no.~3, pp. 1086--1095, 2019.

\bibitem{gupta2022unpacking}
A.~Gupta, A.~Pacchiano, Y.~Zhai, S.~Kakade, and S.~Levine, ``Unpacking reward shaping: Understanding the benefits of reward engineering on sample complexity,'' \emph{Advances in Neural Information Processing Systems}, vol.~35, pp. 15\,281--15\,295, 2022.

\bibitem{kober2013reinforcement}
J.~Kober, J.~A. Bagnell, and J.~Peters, ``Reinforcement learning in robotics: A survey,'' \emph{The International Journal of Robotics Research}, vol.~32, no.~11, pp. 1238--1274, 2013.

\bibitem{arulkumaran2017deep}
K.~Arulkumaran, M.~P. Deisenroth, M.~Brundage, and A.~A. Bharath, ``Deep reinforcement learning: A brief survey,'' \emph{IEEE Signal Processing Magazine}, vol.~34, no.~6, pp. 26--38, 2017.

\bibitem{singh2022reinforcement}
B.~Singh, R.~Kumar, and V.~P. Singh, ``Reinforcement learning in robotic applications: a comprehensive survey,'' \emph{Artificial Intelligence Review}, vol.~55, no.~2, pp. 945--990, 2022.

\bibitem{eschmann2021reward}
J.~Eschmann, ``Reward function design in reinforcement learning,'' \emph{Reinforcement learning algorithms: Analysis and Applications}, pp. 25--33, 2021.

\bibitem{ibrahim2024comprehensive}
S.~Ibrahim, M.~Mostafa, A.~Jnadi, H.~Salloum, and P.~Osinenko, ``Comprehensive overview of reward engineering and shaping in advancing reinforcement learning applications,'' \emph{IEEE Access}, 2024.

\bibitem{ng2000algorithms}
A.~Y. Ng, S.~Russell \emph{et~al.}, ``Algorithms for inverse reinforcement learning.'' in \emph{Icml}, vol.~1, no.~2, 2000, p.~2.

\bibitem{abbeel2004apprenticeship}
P.~Abbeel and A.~Y. Ng, ``Apprenticeship learning via inverse reinforcement learning,'' in \emph{Proceedings of the twenty-first international conference on Machine learning}, 2004, p.~1.

\bibitem{ratliff2006maximum}
N.~D. Ratliff, J.~A. Bagnell, and M.~A. Zinkevich, ``Maximum margin planning,'' in \emph{Proceedings of the 23rd international conference on Machine learning}, 2006, pp. 729--736.

\bibitem{syed2007game}
U.~Syed and R.~E. Schapire, ``A game-theoretic approach to apprenticeship learning,'' \emph{Advances in neural information processing systems}, vol.~20, 2007.

\bibitem{ziebart2008maximum}
B.~D. Ziebart, A.~L. Maas, J.~A. Bagnell, A.~K. Dey \emph{et~al.}, ``Maximum entropy inverse reinforcement learning.'' in \emph{Aaai}, vol.~8.\hskip 1em plus 0.5em minus 0.4em\relax Chicago, IL, USA, 2008, pp. 1433--1438.

\bibitem{levine2018learning}
S.~Levine, P.~Pastor, A.~Krizhevsky, J.~Ibarz, and D.~Quillen, ``Learning hand-eye coordination for robotic grasping with deep learning and large-scale data collection,'' \emph{The International journal of robotics research}, vol.~37, no. 4-5, pp. 421--436, 2018.

\bibitem{shah2022inverse}
S.~I.~H. Shah, A.~Coronato, and M.~Naeem, ``Inverse reinforcement learning based approach for investigating optimal dynamic treatment regime,'' in \emph{Workshops at 18th International Conference on Intelligent Environments (IE2022)}.\hskip 1em plus 0.5em minus 0.4em\relax IOS Press, 2022, pp. 266--276.

\bibitem{uchibe2018model}
E.~Uchibe, ``Model-free deep inverse reinforcement learning by logistic regression,'' \emph{Neural Processing Letters}, vol.~47, no.~3, pp. 891--905, 2018.

\bibitem{codevilla2018end}
F.~Codevilla, M.~M{\"u}ller, A.~L{\'o}pez, V.~Koltun, and A.~Dosovitskiy, ``End-to-end driving via conditional imitation learning,'' in \emph{2018 IEEE international conference on robotics and automation (ICRA)}.\hskip 1em plus 0.5em minus 0.4em\relax IEEE, 2018, pp. 4693--4700.

\bibitem{davila2013environmental}
A.~Davila, E.~Del~Pozo, E.~Aramburu, and A.~Freixas, ``Environmental benefits of vehicle platooning,'' SAE Technical Paper, Tech. Rep., 2013.

\bibitem{bishop2017evaluation}
R.~Bishop, D.~Bevly, L.~Humphreys, S.~Boyd, and D.~Murray, ``Evaluation and testing of driver-assistive truck platooning: Phase 2 final results,'' \emph{Transportation research record}, vol. 2615, no.~1, pp. 11--18, 2017.

\bibitem{bai2022approximate}
T.~Bai, A.~Johansson, K.~H. Johansson, and J.~M{\aa}rtensson, ``Approximate dynamic programming for platoon coordination under hours-of-service regulations,'' in \emph{2022 IEEE 61st Conference on Decision and Control (CDC)}.\hskip 1em plus 0.5em minus 0.4em\relax IEEE, 2022, pp. 7663--7669.

\bibitem{wei2024multi}
D.~Wei, P.~Yi, J.~Lei, and X.~Zhu, ``Multi-agent deep reinforcement learning for distributed and autonomous platoon coordination via speed-regulation over large-scale transportation networks,'' \emph{arXiv preprint arXiv:2412.01075}, 2024.

\bibitem{oliehoek2016concise}
F.~A. Oliehoek, C.~Amato \emph{et~al.}, \emph{A concise introduction to decentralized POMDPs}.\hskip 1em plus 0.5em minus 0.4em\relax Springer, 2016, vol.~1.

\bibitem{chang2024survey}
Y.~Chang, X.~Wang, J.~Wang, Y.~Wu, L.~Yang, K.~Zhu, H.~Chen, X.~Yi, C.~Wang, Y.~Wang \emph{et~al.}, ``A survey on evaluation of large language models,'' \emph{ACM transactions on intelligent systems and technology}, vol.~15, no.~3, pp. 1--45, 2024.

\end{thebibliography}

\end{document}